\documentclass[%
 reprint,
 amsmath,amssymb,
 aps,
]{revtex4-2}
\usepackage{dblfloatfix}
\usepackage{float} % In the preamble
\usepackage{placeins}
\usepackage{graphicx}% Include figure files
\usepackage{dcolumn}% Align table columns on decimal point
\usepackage{bm}% bold math
\usepackage{hyperref}% add hypertext capabilities
\begin{document}

\preprint{APS/123-QED}

\title{Determination of Particle-Size Distributions from Light-Scattering Measurement Using Constrained Gaussian Process Regression}

\author{Fahime Seyedheydari}
\author{Mahdi Nasiri}
\author{Marcin Mińkowski}
\author{Simo Särkkä}
\affiliation{Aalto University, Department of Electrical Engineering and Automation (EEA), Espoo, Finland}

\begin{abstract}
In this work, we propose a novel methodology for robustly estimating particle size distributions from optical scattering measurements using constrained Gaussian process regression. The estimation of particle size distributions is commonly formulated as a Fredholm integral equation of the first kind, an ill-posed inverse problem characterized by instability due to measurement noise and limited data. To address this, we use a Gaussian process prior to regularize the solution and integrate a normalization constraint into the Gaussian process via two approaches: by constraining the Gaussian process using a pseudo-measurement and by using Lagrange multipliers in the equivalent optimization problem. To improve computational efficiency, we employ a spectral expansion of the covariance kernel using eigenfunctions of the Laplace operator, resulting in a computationally tractable low-rank representation without sacrificing accuracy. Additionally, we investigate two complementary strategies for hyperparameter estimation: a data-driven approach based on maximizing the unconstrained log marginal likelihood, and an alternative approach where the physical constraints are taken into account. Numerical experiments demonstrate that the proposed constrained Gaussian process regression framework accurately reconstructs particle size distributions, producing numerically stable, smooth, and physically interpretable results. This methodology provides a principled and efficient solution for addressing inverse scattering problems and related ill-posed integral equations.
\end{abstract}


\maketitle

%\tableofcontents

\section{Introduction}

The characterization of particle size distributions (PSDs) plays a crucial role in understanding and controlling the physical properties of particulate systems across a wide range of scientific and engineering applications~\cite{allen2013particle}. Accurate knowledge of PSDs is essential in fields such as atmospheric science~\cite{willeke1975atmospheric}, pharmaceuticals~\cite{shekunov2007particle}, and materials processing~\cite{gupta1998fluid}, where particle size and distribution directly influence optical, mechanical, and chemical properties~\cite{bohren2008absorption, mishchenko2002scattering}.

The light scattering response of particles is strongly influenced by their size distribution~\cite{hulst1981light}. Optical spectroscopy is widely used to determine PSDs due to its non-destructive nature and its ability to provide rapid and reliable measurements across a broad size range~\cite{bohren2008absorption}. Accurate estimation of size distributions is essential for both modeling the scattering behavior of particle ensembles~\cite{mishchenko2002scattering} and interpreting optical measurements with precision~\cite{twomey2019introduction}.

In the study of scattering processes, two principal formulations are commonly considered: the direct problem and the inverse problem~\cite{colton1998inverse}. The direct problem involves computing the scattered field from a known physical system, such as the particle composition and size~\cite{hulst1981light}. In contrast, the inverse problem aims to infer unknown physical characteristics, such as PSD, from measured scattering data, whether obtained through laboratory experiments or remote sensing~\cite{twomey2019introduction}.

The relationship between the measured scattering data and the PSD is commonly expressed as a Fredholm integral equation of the first kind~\cite{twomey2019introduction}. This formulation relates the observed scattering data to the unknown size distribution through an integral operator. The Fredholm integral equation technique was originally introduced in the context of light scattering by Holt et al.~\cite{holt1978integral}. Early applications within this framework included the retrieval of heterogeneous aerosol size distributions~\cite{twomey1963measurements,yamamoto1969determination}.

However, the inverse scattering problem is inherently ill-posed, meaning that small perturbations in the data can lead to large variations in the solution. According to Hadamard’s definition~\cite{john1991partial}, a well-posed problem must satisfy three key conditions: (1) existence of a solution, (2) uniqueness of the solution, and (3) stability of the solution with respect to small changes in the input. Many inverse problems, including those involving light scattering, violate one or more of these conditions~\cite{tikhonov1977solutions}.

The estimation of particle size from light scattering represents a classic inverse problem in optical dispersion systems~\cite{twomey2019introduction,bohren2008absorption}. This method constitutes a mathematical framework for structuring light scattering in the form of an integral equation,
\begin{equation}
\mu(\lambda) = \int_{\Omega_1} A (\lambda, r) \rho(r) \, dr + \epsilon(\lambda), \quad \lambda \in \Omega_2,
\label{eq:Fredholm1}
\end{equation}
where \( \mu(\lambda) \) represents the measured scattering data at wavelength \(\lambda\), \( \epsilon(\lambda) \) is zero-mean white noise that models measurement uncertainty, \( A(\lambda, r) \) is the known kernel describing how particles of size \(r\) scatter light at wavelength \(\lambda\), and \(\rho(r)\) is the unknown PSD to be estimated. The domains \( \Omega_1 \subseteq \mathbb{R}^{d_1}\) and \(\Omega_2 \subseteq \mathbb{R}^{d_2}\) correspond to the range of particle sizes and wavelengths, respectively, where \(d_1\) and \(d_2\) in our case are both 1.

In this work, we formulate an inverse problem aimed at reconstructing the distribution of a continuous parameter \(\rho(r)\) from measured scattering data \(\mu(\lambda)\). The underlying integral relationship is modeled by a Fredholm integral equation of the first kind, which is inherently ill-posed~\cite{phillips1962technique, tikhonov1963solution, engl1996regularization}. To address the ill-posed nature of the inverse problem, we employ regularization techniques that stabilize the solution and suppress noise amplification.

To solve this inverse problem, we employ Gaussian process regression (GPR), a nonparametric Bayesian method that provides a principled framework for regularization~\cite{rasmussen2006gaussian, bishop2006pattern}. We demonstrate a connection between classical regularization and GPR, leveraging it as a statistically grounded machine learning framework for solving the integral equation. GPR defines a distribution over functions by placing a Gaussian prior over \(\rho(r)\) and a Gaussian likelihood based on the measurement model. This formulation leads to a regularized least-squares problem, where the trade-off between data fit and prior smoothness is defined by the covariance structure of the GP.

Constrained GPR integrates domain knowledge into regression models via hard constraints to improve predictive performance when data are limited. Several classes of constraints can be imposed on GP models~\cite{swiler2020survey}, including boundary conditions and value constraints~\cite{solin2019know, solin2020hilbert, graepel2003solving}, monotonicity and convexity constraints~\cite{kelly1990monotone, maatouk2023finite, riihimaki2010gaussian}, bound or positivity constraints~\cite{maatouk2017gaussian, jensen2013bounded, hertzberg2018chemical, da2012gaussian}, and linear differential equation constraints~\cite{raissi2017machine, graepel2003solving, sarkka2011linear}. These constraints can be incorporated into various components of GP models, such as the posterior prediction of the latent function~\cite{jensen2013bounded, da2012gaussian}, the likelihood of observations~\cite{jensen2013bounded}, the covariance kernels~\cite{raissi2017machine, sarkka2011linear, jidling2017linearly}, or the input data used in the regression model~\cite{salzmann2010implicitly}.

In addition to the aforementioned approach of using constrained GPR, we also study an alternative formulation of the problem as a constrained optimization problem~\cite{swiler2020survey}, where a Lagrange multiplier explicitly enforces the integral constraint within the regularized least-squares objective. Furthermore, to enable practical computation, we perform a spectral expansion of the covariance kernel~\cite{solin2020hilbert, purisha2019probabilistic}, providing a rigorous foundation for discretizing the problem. This transformation allows the infinite-dimensional formulation to be approximated in a finite-dimensional subspace while preserving the essential smoothness properties of the prior and the constraint.

The GPR formulation of the problem also enables us to estimate the model's hyperparameters using a maximum (marginal) likelihood approach. In addition to the standard marginal likelihood approach based solely on the observed scattering data, we also consider a joint marginal likelihood formulation for hyperparameter estimation. This joint approach integrates both the measurement model and the normalization constraint, enabling physically consistent hyperparameter inference that captures the balance between the fit to the data and prior knowledge.

This work presents a structured and physically consistent framework for estimating PSDs from optical scattering data by solving an ill-posed inverse problem. Section~\ref{sec:Methods} outlines the overall methodological approach. We begin in Section~\ref{sec:Lorenz-Mie} by applying Lorenz-Mie theory to define the scattering behavior of polydisperse systems. The inverse problem is formulated as a Fredholm integral equation of the first kind, as described in Section~\ref{sec:GPR}. To address its inherent ill-posedness, we employ GPR, a nonparametric Bayesian method that enables flexible regularization and uncertainty quantification. Section~\ref{sec:Cov_fun} discusses the role of covariance functions in encoding prior smoothness assumptions, while Section~\ref{sec:Cla_Reg} establishes a connection between GPR and classical regularization theory. To improve computational efficiency, Section~\ref{sec:Ker_Eig} introduces a kernel eigenfunction expansion that reduces the complexity of the problem. The inclusion of a normalization constraint via Lagrange multipliers, detailed in Sections~\ref{sec:Constrain_OPt} and~\ref{sec:For_Nor}, ensures physical consistency of the estimated distribution. The posterior solution under both measurement data and constraints is efficiently derived within the eigenfunction framework, as described in Section~\ref{sec:Ker_Ei_Ex}. Section~\ref{sec:Param_Est} then presents strategies for hyperparameter estimation, including both a standard log marginal likelihood approach (Section~\ref{sec:Log_Mar}) and a joint formulation that incorporates the constraint (Section~\ref{sec:Cons_only}). In Section~\ref{sec:Result}, we demonstrate the accuracy and robustness of the proposed framework through numerical experiments. Section~\ref{sec:Conclu} concludes with a summary of our key findings and emphasizes the stability, flexibility, and physical consistency of the method.

\section{\label{sec:Methods}Methods}
In this section, we present a comprehensive framework for estimating PSDs from light-scattering measurements by solving an ill-posed inverse problem. The estimation is formulated as a Fredholm integral equation of the first kind, in which the scattering behavior of particles is described using Lorenz-Mie theory~\cite{mie1976contributions,hulst1981light}. To address the inherent instability of the inverse problem, we employ GPR, which introduces regularization through prior covariance functions and ensures smooth and stable solutions. Additionally, we incorporate a normalization constraint using Lagrange multipliers to maintain the physical validity of the estimated distributions. To enhance computational efficiency, we apply a kernel eigenfunction expansion that reduces the complexity of the problem. 

\subsection{\label{sec:Lorenz-Mie}Lorenz-Mie Theory}
In analyzing dispersive media composed of uniformly shaped materials with different sizes, it is essential to define ensemble averages for particle size in relation to the scattering coefficient. The scattering coefficient attributed to Mie scattering, denoted as $\mu_{\textrm{sca}}$, can be determined from scattering measurements acquired across a range of wavelengths~\cite{mie1976contributions} and is given by
\begin{equation}
\mu_{\rm{sca}}(\lambda)= \int_{\Omega_1} \frac{3}{4} \dfrac{fQ_{\rm{sca}}(\lambda, r)}{r} \rho(r) \, dr,
\label{eq:Fredholm}
\end{equation}
where \(Q_{\rm{sca}}(\lambda, r)\) is the scattering efficiency of a particle of radius \(r\) at wavelength \(\lambda\). In this context, \(f\) and \(\rho(r)\) are the volume fraction and the distribution of particle size, respectively. All of the distributions are normalized to unity such that,
\begin{equation}
\int_{\Omega_1} \rho(r) \, dr = 1.
\label{eq:Distri}
\end{equation}

The Lorenz-Mie theory provides a foundational approach for analyzing the interaction between spherical particles and plane electromagnetic waves~\cite{mie1976contributions,seyedheydari2022electromagnetic}. This theory yields exact analytical solutions to Maxwell's equations for the interaction of electromagnetic radiation with homogeneous spherical particles, thereby providing a precise determination of their optical properties, including scattering, absorption, and extinction efficiencies. When light strikes an isotropic, homogeneous spherical particle, scattering efficiency, denoted as $Q_\text{sca}$, is computed using Lorenz-Mie theory~\cite{hulst1981light} as follows:
\begin{equation}
Q_\text{sca}=\frac{2}{y^2}\sum_{n=1}^{N} (2n+1)(|a_n|^2+|b_n|^2),
\end{equation}
where \(y = \frac{2\pi n_{m} R}{\lambda}\) is a size parameter, \(R\) is the particle radius, and $n_m$ is the refractive index of the surrounding medium. Above, $a_n$ and $b_n$ are the Mie coefficients. The order of the dielectric resonance is denoted by $n$, with \(n=1\) corresponding to the dipole mode, \(n=2\) to the quadrupole, \(n=3\) to the octupole, and so forth~\cite{bohren2008absorption}. The summation index \(N\) defines the number of multipole terms retained in the Mie series expansion, typically selected based on the size parameter \(y\) to ensure numerical convergence. A commonly used empirical approximation is
\(N \approx y + 4y^{1/3} + 2\) \cite{bohren2008absorption}.

In practical scenarios, the scattering coefficient, $\mu_{\textrm{sca}}$, often describes media that consist of mixtures of particles with varying sizes, shapes, and refractive indices. This study addresses the inverse problem of determining the particle size distribution in polydisperse media, where the constituent spheres exhibit varying sizes.

Under the assumption that scattering events occur independently within the medium, the transformation matrix representing the optical properties for a unit volume can be described using Eq.~\eqref{eq:Fredholm}. By defining
\begin{equation}
A(\lambda, r) = \frac{3}{4} \dfrac{fQ_{\rm{sca}}(\lambda, r)}{r},
\label{eq:A=Q}
\end{equation}
the relationship between the scattering coefficient and its corresponding scattering efficiency spectrum is expressed as a Fredholm integral equation of the first kind in Eq.~\eqref{eq:Fredholm1}. A Fredholm integral equation of the first kind is non-singular if the integration range is finite and the kernel, \( A(\lambda, r) \), is bounded; otherwise, it is singular~\cite{phillips1962technique}.

A common method for solving integral equations is to represent them as operator equations within a Hilbert space framework. In this context, the bounded linear operator \(\mathcal{A} : L^2(\Omega_1) \rightarrow L^2(\Omega_2)\) is defined as
\begin{equation}
(\mathcal{A}\rho)(\lambda) = \int_{\Omega_1} A(\lambda, r) \rho(r) \, dr, \quad \lambda \in \Omega_2,
\end{equation}
where \(\mathcal{A}\) is a linear integral operator that maps the unknown function \(\rho(r)\) (PSD) to the observed function \(\mu(\lambda)\), and this can be written concisely as:
\begin{equation}
\mu = \mathcal{A} \rho +\epsilon,
\label{linear}
\end{equation}
where \(\epsilon\) represents measurement noise. The integral operator \(\mathcal{A}\) represents a continuous transformation that is sometimes called a Hilbert-Schmidt integral operator~\cite{vogel2002computational}. 

\subsection{\label{sec:GPR}Gaussian Process Regression}

A GP can be viewed as a generalization of the multivariate Gaussian distribution to infinite-dimensional function spaces. It provides a principled, probabilistic framework for modeling functions and is fully specified by a mean function \(m(r)\) and a covariance function \( k(r, r') \)~\cite{rasmussen2006gaussian}. A GP defines a distribution over functions such that, for any finite collection of input points 
\(r_1, r_2, \ldots,r_N\), the corresponding random vector \( \rho = [\rho (r_1), \rho (r_2), \ldots, \rho (r_N)]^\intercal\) follows a multivariate normal distribution. This is denoted in compact form as:
\begin{equation}
\rho(r) \sim \mathcal{GP}(m (r), k(r, r')).
\label{eq:GP}
\end{equation}

In practical applications, the mean function is often assumed to be zero, \( m(r) = 0 \), unless prior knowledge suggests otherwise. This assumption does not limit the expressiveness of the model, as the posterior mean can vary flexibly after observing data~\cite{rasmussen2006gaussian,sarkka2023bayesian}.

To connect the latent function \(\rho (r)\) with observable data, we introduce a linear measurement model that maps \(\rho (r)\) to a set of observations \(\mu_i\) through a known operator \(\mathcal{A}\) and additive Gaussian noise:
\begin{align}
\rho(r)& \sim \mathcal{GP}(0, k(r, r')), \notag \\
\mu_i & = (\mathcal{A} {\rho})(\lambda_i) + {\epsilon}_i, \quad \epsilon_i \sim \mathcal{N}(0, \sigma^2 I),
\label{eq:measurment}
\end{align}
where \(\mathcal{A}\) is a linear operator acting on the function \({\rho}\). The noise term \(\epsilon\) follows a normal distribution, and \(\sigma^2 I\) represents the noise covariance matrix.

Given the joint Gaussianity of the latent function values and the measurements, the posterior distribution can be derived using the standard conditioning rule for multivariate Gaussian distributions. If $\mathbf{a}$ and $\mathbf{b}$ are jointly Gaussian:
\begin{equation}
\begin{bmatrix}
\mathbf{a} \\
\mathbf{b}
\end{bmatrix}
\sim \mathcal{N}\left(
\begin{bmatrix}
\boldsymbol{\mathcal{M}}_a \\
\boldsymbol{\mathcal{M}}_b
\end{bmatrix},
\begin{bmatrix}
\Sigma_{aa} & \Sigma_{ab} \\
\Sigma_{ba} & \Sigma_{bb}
\end{bmatrix}
\right),
\label{eq:cond-dist}
\end{equation}
then the conditional distribution of $\mathbf{b}$ given $\mathbf{a}$ is expressed as:
\begin{equation}
\mathbf{b} | \mathbf{a} \sim \mathcal{N}\left(\boldsymbol{\mathcal{M}}_b + \Sigma_{ba} \Sigma_{aa}^{-1} (\mathbf{a} - \boldsymbol{\mathcal{M}}_a), \Sigma_{bb} - \Sigma_{ba} \Sigma_{aa}^{-1} \Sigma_{ab}\right).
\label{eq:cond-dist1}
\end{equation}

We aim to estimate the value of the latent function $\rho(r)$ at a test location $r^*$, denoted by $\rho(r^*)$, given the observed data $\mu$. Under the GP prior and the linear measurement model, the joint distribution of the observed data $\mu$ and the test output $\rho(r^*)$ is itself Gaussian. Specifically, the joint distribution is given by a zero-mean Gaussian with the covariance structure determined by the kernel function $k(r, r')$ and the linear operator $\mathcal{A}$. The resulting joint distribution is:
\begin{equation}
\begin{bmatrix}
\mu \\
\rho(r^*) 
\end{bmatrix}
\sim \mathcal{N}\left(
\begin{bmatrix}
0 \\
0
\end{bmatrix},
\begin{bmatrix}
\mathcal{A} k \mathcal{A}^* +\sigma^2 I & \mathcal{A} k(\cdot, r^*) \\
k(r^*, \cdot) \mathcal{A}^*& k(r^*, r^*)
\end{bmatrix}
\right),
\end{equation}
where \(k(\cdot, r^*)\) denotes the covariance between the function values at the training points and the test point $r^*$, while \(k(r^*, r^*) \) represents the prior variance of $\rho(r)$ at the test point. The expression $\mathcal{A}k(\cdot, r^*)$ is the covariance between the observed data and the function value at the test point. The expression $\mathcal{A}k\mathcal{A}^*$ denotes applying the operator $\mathcal{A}$ to $k$ from the left and $\mathcal{A}^*$ from the right, and it gives the covariance of the measurement data under the GP prior (cf.\ \cite{sarkka2011linear}).

Thus, applying the conditional distribution formula for jointly Gaussian random variables, the posterior mean and covariance of the function value at an unseen test location \(r^*\), denoted by \(\rho(r^*)\), are given by:
\begin{align}
\mathbb{E}[\rho(r^*)] &= k(r^*, \cdot)\mathcal{A}^* \left(\mathcal{A} k \mathcal{A}^* + \sigma^2 I\right)^{-1} \mu, \label{eq:test_gp_mean} \\
\mathrm{Cov}[\rho(r^*)] &= k(r^*, r^*) \notag\\
&\quad - k(r^*, \cdot) \mathcal{A}^* \left(\mathcal{A} k \mathcal{A}^* + \sigma^2 I\right)^{-1} \mathcal{A} k(\cdot, r^*),
\end{align}
where \(k_* = k(\cdot, r^*)\) is the cross-covariance vector between the training points and the test point.

\subsection{\label{sec:Cov_fun}Covariance Function}
The covariance function \( k(r, r')\) plays a crucial role in encoding structural properties of the function \(\rho \), such as smoothness, periodicity, and correlation length. For stationary covariance functions, the covariance can be analyzed using its spectral density—the Fourier transform of the covariance function~\cite{rasmussen2006gaussian},
\begin{equation}
S(\boldsymbol{\omega}) = \int k(\boldsymbol{\mathrm{r}}) e^{-i\boldsymbol{\omega}^\intercal\boldsymbol{\mathrm{r}}} d\boldsymbol{\mathrm{r}}.
\label{eq:kernel}
\end{equation}
Common choices for \( k(r, r')\) include the squared exponential (SE) kernel, the Matérn kernel, and periodic kernels, each of which imposes different assumptions on the behavior of \( \rho \). The SE kernel is defined as~\cite{rasmussen2006gaussian}
\begin{equation}
k_{\text{SE}}(\boldsymbol{\mathrm{r}}) = \sigma_f^2 \exp\left(-\frac{\|\boldsymbol{\mathrm{r}}\|_2^2}{2\ell^2}\right),
\label{eq:kernel12}
\end{equation}
where \(\sigma_f ^2\) is the signal variance, and \( \ell \) is the length-scale hyperparameter. 
The Fourier transform of the SE kernel yields the following spectral density:
\begin{equation}
S_{\text{SE}}(\boldsymbol{\omega}) = \sigma_f^2\sqrt{2\pi}\ell\exp\left(-\frac{\ell^2\|\boldsymbol{\omega}\|_2^2}{2}\right).
\label{eq:kerc}
\end{equation}

In this work, we also consider the Mat\'ern covariance function as a prior in our GP model~\cite{rasmussen2006gaussian}. In one dimension, the Mat\'ern covariance function is defined as
\begin{equation}
k_{\text{Mat\'ern}}(\boldsymbol{\mathrm{r}}) = \sigma_f^2 \frac{2^{1 - \nu}}{\Gamma(\nu)} \left(\frac{ \sqrt{2\nu} \|\boldsymbol{\mathrm{r}}\|_2}{\ell} \right)^{\nu} K_{\nu} \left( \frac{\sqrt{2\nu}\|\boldsymbol{\mathrm{r}}\|_2}{\ell} \right),
\end{equation}
where $\nu$ controls the smoothness, and $K_{\nu}$ is the modified Bessel function of the second kind. The Mat\'ern covariance is particularly flexible, as it allows control over the smoothness of the reconstructed function through the parameter $\nu$. The spectral density corresponding to the Mat\'ern covariance function is given by
\begin{align}
S_{\text{Mat\'ern}}(\boldsymbol{\omega}) =\; & \sigma_f^2 \frac{2 \sqrt{\pi} \Gamma\left( \nu + 1/2 \right) (2\nu)^{\nu}}{\Gamma(\nu)\; \ell^{2\nu}} \notag \\
& \times \left( \frac{2\nu}{\ell^2} + \| \boldsymbol{\omega} \|_2^2 \right)^{-\left( \nu + 1/2 \right)}.
\end{align}
In the experiments, we compare the performance of the Mat\'ern covariance with the SE covariance function in terms of reconstruction accuracy and flexibility. The Mat\'ern class provides a more general framework, including less smooth and more realistic prior assumptions by adjusting $\nu$. This comparison allows us to evaluate the impact of prior smoothness assumptions on the quality of the reconstruction.

\subsection{\label{sec:Cla_Reg}Classical Regularization}

The inverse problem of estimating \(\rho(r)\) from noisy measurements \(\mu(\lambda)\) is classically formulated as a least-squares problem~\cite{tikhonov1977solutions}. To address the ill-posed nature of this problem and suppress instability due to measurement noise, a classical regularization method, known as Tikhonov regularization~\cite{tikhonov1963solution}, can be applied. This leads to the following regularized least-squares formulation:
\begin{equation}
\mathcal{J}[\rho] = \frac{1}{2\sigma^2} \sum_{i=1}^n \left( \mu_i - (\mathcal{A} \rho)(\lambda_i) \right)^2 + \frac{1}{2\sigma_f^2} \int_{\Omega_1} |\mathcal{R} \rho(r)|^2 \, dr,
\label{eq:Tikhon}
\end{equation}
where \( \mathcal{R} \) is a linear regularization operator. The term \(\sigma^2\) denotes the noise variance associated with the measurements, while \(\sigma_f^2\) is a regularization coefficient. However, it turns out that, with a suitable choice of $\mathcal{R}$, minimizing this functional is equivalent to computing the posterior mean of the GPR model in Eq.~\eqref{eq:test_gp_mean} (see, e.g., \cite{purisha2019probabilistic}).

\subsection{\label{sec:Ker_Eig}Kernel Eigenfunction Expansion} 

For practical implementation, we approximate the integral in Eq.~\eqref{eq:Fredholm} using numerical discretization, converting the continuous problem into a computationally tractable form. Here, we adopt a GP prior over the particle size distribution \(\rho(r)\), defined by a covariance function \(k(r, r')\) that encodes assumptions about smoothness and structure through its spectral density. To reduce computational complexity and enable efficient GP inference, we employ a low-rank approximation based on eigenfunction expansions, as introduced by Solin and S\"arkk\"a~\cite{solin2020hilbert}. Specifically, the covariance kernel is represented by a truncated sum of basis functions:
\begin{equation}
k(r, r') \approx \sum_{j=1}^q S(\sqrt{\lambda_j}) \phi_j(r) \phi_j(r'),
\end{equation}
where \(S(\cdot)\) denotes the spectral density and \(\phi_j(r)\) are eigenfunctions of the Laplace operator:
\begin{equation}
\phi_j(r) = \sqrt{\frac{1}{L}} \sin\left(\frac{j\pi (r + L)}{2L}\right), \quad \lambda_j = \left(\frac{j \pi}{2L}\right)^2,
\end{equation}
with domain size \(L = r_{\text{max}} - r_{\text{min}}\). Here, \(q\) denotes the number of basis functions used in the truncated expansion.

Rather than solving directly for \(\rho(r)\), we approximate it using this truncated basis expansion:
\begin{equation}
\rho(r)= \sum_{j=1}^q \alpha_j \phi_j(r), \quad \alpha_j \sim \mathcal{N}(0, S(\sqrt{\lambda_j})).
\label{eq:rho88}
\end{equation}
Letting \(\Phi\) be the \(n \times q\) matrix of eigenfunctions evaluated at \(n\) discrete points, the prior distribution of \(\alpha = [\alpha_1, \dots, \alpha_q]^\intercal\) becomes:
\begin{equation}
\alpha \sim \mathcal{N}(0, \Lambda), \quad \Lambda = \text{diag}(S(\sqrt{\lambda_1}), \dots, S(\sqrt{\lambda_q})).
\end{equation}
Substituting the eigenbasis expansion into the forward scattering model, each measurement \(\mu(\lambda_i)\) is approximated by:
\begin{equation}
\mu(\lambda_i) \approx \sum_{j=1}^{q} \alpha_j \int_{\Omega_1} A(\lambda_i, r) \phi_j(r) \, dr.
\end{equation}
To facilitate numerical computations, we define the projection matrix:
\begin{equation}
[\Psi_\mathcal{A}]_{ij} = \sum_k^n A(\lambda_i, r_k) \phi_j(r_k) \Delta r_k,
\end{equation}
yielding the compact linear system:

\begin{equation}
\mu = \Psi_\mathcal{A}\alpha + \epsilon.
\label{eq:model}
\end{equation}

We first represent the particle size distribution $\rho(r)$ using a truncated expansion in terms of a set of orthonormal eigenfunctions of the Laplace operator, as defined in Eq.~\eqref{eq:rho88}. Substituting this expansion into the forward scattering model yields a linear relationship between the measurements and the coefficient vector $\alpha$, as expressed in Eq.~\eqref{eq:model}. 

We can also convert this into a classical type of optimization problem, which we discussed in Section~\ref{sec:Cla_Reg}. The likelihood term gets converted into $\frac{1}{2\sigma^2} \|\mu - \Psi_A \alpha\|^2$ resulting from the assumed Gaussian noise model, while the prior term becomes $\frac{1}{2} \alpha^\top \Lambda^{-1} \alpha$ from the GP prior on $\rho(r)$ with the diagonal covariance matrix $\Lambda$. The combination of these terms leads directly to the regularized least-squares objective function,
\begin{equation}
\mathcal{J}(\alpha) = \frac{1}{2\sigma^2} \|\mu - \Psi_\mathcal{A} \alpha\|^2 + \frac{1}{2} \alpha^\intercal \Lambda^{-1} \alpha,
\label{eq:matrix}
\end{equation}
which is a basis function expansion based approximation to the cost functional in Eq.~\eqref{eq:Tikhon}.

To enable Bayesian inference of the particle size distribution, we constructed a joint probabilistic model within the GP framework. Specifically, the measured scattering data $\mu$ and the unknown distribution $\rho$ were modeled jointly as a multivariate Gaussian distribution. Employing eigenfunction expansions characterized by the matrices $\Psi_{\mathcal{A}}$ and $\Phi$, and incorporating the covariance structure described by the spectral matrix $\Lambda$, the joint Gaussian distribution is explicitly represented as follows:
\begin{equation}
\begin{bmatrix}
\mu \\
\rho
\end{bmatrix}
\sim \mathcal{N}\left(
0,
\begin{bmatrix}
\Psi_\mathcal{A}\Lambda \Psi_\mathcal{A}^\intercal + \sigma^2 I & \Psi_\mathcal{A} \Lambda \Phi^\intercal \\
\Phi \Lambda \Psi_\mathcal{A}^\intercal & \Phi \Lambda \Phi^\intercal
\end{bmatrix}
\right).
\end{equation}

Applying GPR, the posterior mean and covariance of \(\rho\) given the measurements \(\mu\) are approximated as:
\begin{align}
\mathbb{E}[\rho \mid \mu] &\approx \Phi \Lambda \Psi_\mathcal{A}^\intercal \left( \Psi_\mathcal{A} \Lambda \Psi_\mathcal{A}^\intercal + \sigma^2 I \right)^{-1} \mu, \label{eq:posterior_mean} \\
\text{Cov}[\rho \mid \mu] &\approx \Phi \Lambda \Phi^\intercal 
- \Phi \Lambda \Psi_\mathcal{A}^\intercal \left( \Psi_\mathcal{A} \Lambda \Psi_\mathcal{A}^\intercal + \sigma^2 I \right)^{-1} \Psi_\mathcal{A} \Lambda \Phi^\intercal. \label{eq:posterior_cov}
\end{align}
To efficiently address the computational challenges associated with large matrix inversions—common in high-dimensional inverse problems—we apply the Woodbury matrix identity. This reformulation yields computationally efficient expressions for the posterior mean and covariance:
\begin{align}
\mathbb{E}[\rho \mid \mu] &\approx \Phi \left( \Psi_\mathcal{A}^\intercal \Psi_\mathcal{A} + \sigma^2 \Lambda^{-1} \right)^{-1} \Psi_\mathcal{A}^\intercal \mu, \label{eq:woodbury_mean} \\
\text{Cov}[\rho \mid \mu] &\approx \Phi \left( \frac{1}{\sigma^2} \Psi_\mathcal{A}^\intercal \Psi_\mathcal{A} + \Lambda^{-1} \right)^{-1} \Phi^\intercal. \label{eq:woodbury_cov}
\end{align}
This revised formulation significantly reduces computational complexity while maintaining accuracy, making it practical for large-scale inverse scattering problems.

\subsection{\label{sec:Constrain_OPt}Constrained Optimization using Lagrange Multipliers} 
Estimating the unknown distribution \( \rho \) from noisy observational data \( \mu \) is a challenging inverse problem, often formulated as a regularized optimization problem. The aim is to recover a plausible distribution that aligns with the measured data while incorporating prior knowledge about the underlying physical system. In the classical (Tikhonov-based) approach, the regularized solution is obtained by minimizing the cost functional:
\begin{equation}
\hat{\rho}= \arg\min_{\rho} \, \mathcal{J}[\rho],
\label{eq:min}
\end{equation}
where \( \mathcal{J}[\rho] \) is a functional that captures the trade-off between fitting the observed data and incorporating prior knowledge, such as smoothness or regularity. This objective function represents a balance between two components: the first term enforces consistency with the observed data, while the second term acts as a regularization term, encoding prior assumptions about the distribution for more stable and physically plausible predictions. Thus, the overall objective can be expressed as:
\[
\mathcal{J}[\rho] = \mathcal{J}_{\text{data}}(\rho)+ \mathcal{J}_{\text{reg}}(\rho).
\]

In addition to fitting the observed data and incorporating prior information, it is often advantageous to enforce physical or domain-specific constraints to ensure that the estimated distribution remains physically meaningful~\cite{swiler2020survey}. In our case, a critical constraint is the normalization of the distribution \( \rho \), representing the requirement that the total probability sums to unity. This can be formally imposed through a Lagrange multiplier approach, leading to a constrained optimization formulation. Specifically, by introducing a Lagrange multiplier \( c \) to enforce this constraint, the objective functional becomes:
\begin{align}
\mathcal{L}[\rho, c] &= \frac{1}{2\sigma^2} \sum_{i=1}^n \left( \mu_i - (\mathcal{A} \rho)(\lambda_i) \right)^2 + \frac{1}{2\sigma_f^2} \int_{\Omega_1} |\mathcal{R} \rho(r)|^2 \, dr \notag \\
&+ c \left( \int_{\Omega_1} \rho(r) \, dr - 1\right).
\label{full22}
\end{align}
The third term in this formulation explicitly enforces the normalization constraint, ensuring that the total size distribution \( \rho \) sums to unity, as defined in Eq.~\eqref{eq:Distri}. This constraint is essential for maintaining the physical consistency of the estimated distribution, preventing unphysical solutions that might arise from the regularized least-squares formulation in Eq.~\eqref{eq:Tikhon}.

To express this functional in the previously defined matrix form Eq.~\eqref{eq:matrix}, we substitute the finite basis expansion for \(\rho(r)\), resulting in:
\begin{equation}
\mathcal{L}(\alpha, c) = \frac{1}{2\sigma^2}\|\mu - \Psi_\mathcal{A}\alpha\|^2 + \frac{1}{2}\alpha^\intercal \Lambda^{-1}\alpha + c(h^\intercal\alpha - 1),
\end{equation}
where the first two terms correspond to the regularized least-squares objective previously defined, and the third term enforces the normalization condition on \( \rho \) through the vector \( h\), which represents the discrete form of the integral constraint. This cost function represents the maximum a posteriori (MAP) estimation for the coefficients \(\alpha\), combining the data likelihood (first term) and the prior regularization (second term) while enforcing the physical constraint that the total distribution integrates to one (third term). 

With the matrix form of the constrained objective function now defined, the next step is to derive the optimal coefficients \(\alpha\) that minimize this functional. This involves differentiating the Lagrangian with respect to \(\alpha\) and setting the resulting derivative to zero, which leads to the following expression:
\begin{equation}
\left(\frac{1}{\sigma^2}\Psi_\mathcal{A}^\intercal\Psi_\mathcal{A} + \Lambda^{-1}\right)\alpha + c h = \frac{1}{\sigma^2}\Psi_\mathcal{A}^\intercal\mu.
\end{equation}
To simplify the expression and highlight the influence of the regularization and data terms, we define the matrix \(M\) as:
\begin{equation}
M = \frac{1}{\sigma^2}\Psi_\mathcal{A}^\intercal\Psi_\mathcal{A} + \Lambda^{-1}.
\end{equation}
We can now explicitly solve for the Lagrange multiplier \(c\) by enforcing the normalization constraint \(h^\intercal\alpha = 1\), ensuring that the estimated distribution remains physically consistent:
\begin{equation}
c = \frac{\frac{1}{\sigma^2}h^\intercal M^{-1}\Psi_\mathcal{A}^\intercal\mu - 1}{h^\intercal M^{-1} h}.
\end{equation}
Once \(c\) is determined, the optimal solution for the coefficient vector \(\alpha\) can be expressed as:
\begin{equation}
\alpha = M^{-1}\left(\frac{1}{\sigma^2}\Psi_\mathcal{A}^\intercal\mu - c h\right).
\end{equation}
To improve numerical efficiency and stability, the matrix inverse \(M^{-1}\) can be computed using the Woodbury matrix identity,
\begin{equation}
M^{-1} = \Lambda - \Lambda \Psi_\mathcal{A}^\intercal(\sigma^2 I + \Psi_\mathcal{A} \Lambda \Psi_\mathcal{A}^\intercal)^{-1}\Psi_\mathcal{A} \Lambda.
\end{equation}
Finally, the estimate of the distribution \(\rho(r)\) is reconstructed from the basis function expansion as:
\begin{equation}
\hat{\rho}(r) = \Phi(r)\alpha.
\end{equation}
This formulation ensures that measurement noise, prior information, and physical constraints are accurately incorporated into the estimation process, resulting in a solution that is not only stable and computationally efficient but also physically meaningful.

\subsection{\label{sec:For_Nor}Posterior Mean and Covariance}

In this section, the aim is to derive the posterior mean and covariance of the Gaussian process-based solution to the inverse problem with the constraint.

\subsubsection{Posterior Estimation with Direct Operator Representation}
To ensure that the function \(\rho (r)\) represents a physically meaningful size distribution, it must satisfy the integral constraint defined in Eq.~\eqref{eq:Distri}, expressing the requirement that the total distribution sums to unity. This constraint can be expressed using an integral operator \(\mathcal{B}\) that maps the distribution function to a scalar:
\begin{equation}
(\mathcal{B} \rho) = \int_{\Omega_1} \rho(r) dr = 1.
\label{eq:normalization_constraint}
\end{equation}
In this formulation, the operator \(\mathcal{B}\) is a linear map that projects the function \(\rho (r)\) onto a single scalar value, enforcing the normalization constraint within the optimization process.

In the preceding sections, we introduced the measurement model in Eq.~\eqref{eq:measurment} along with the normalization constraint. To form a well-posed estimation of the latent function \(\rho (r)\), both the measurement model and the constraint must be incorporated into a unified probabilistic model. This approach enables the analytical derivation of the posterior distribution using the properties of multivariate Gaussian distributions.

Given the linear nature of both the measurement operator \(\mathcal{A}\) and the constraint operator \(\mathcal{B}\), and assuming a Gaussian prior over \(\rho\), the joint distribution of the observed data and the latent function can be expressed compactly as:
\begin{equation}
y =
\begin{bmatrix}
\mu \\
1
\end{bmatrix}=
\begin{bmatrix}
\mathcal{A} \\
\mathcal{B}
\end{bmatrix}\rho+
\begin{bmatrix}
\epsilon\\
0
\end{bmatrix},
\end{equation}
where the normalization constraint is imposed exactly, without noise. This formulation integrates the measurement model and the constraint within a unified probabilistic framework, allowing inference to be performed through conditioning on the observed data. Accordingly, we express the joint Gaussian distribution over the observations \(y\) and the latent function \(\rho\) as:
\begin{equation}
\begin{bmatrix}
y \\
{\rho}
\end{bmatrix}
\sim \mathcal{N}\left(
\begin{bmatrix}
0 \\
0
\end{bmatrix},
\begin{bmatrix}
\Sigma_{yy} & \Sigma_{y\rho} \\
\Sigma_{\rho y} & \Sigma_{\rho \rho}
\end{bmatrix}
\right),
\label{Sigma_yy}
\end{equation}
where the covariance blocks are defined as follows:
\begin{equation}
\Sigma_{yy} =
\begin{bmatrix}
\mathcal{A} k \mathcal{A}^* + \sigma^2 I & \mathcal{A} k \mathcal{B}^* \\
\mathcal{B} k \mathcal{A}^* & \mathcal{B} k \mathcal{B}^*
\end{bmatrix},
\end{equation}
and the cross-covariance, \(\Sigma_{\rho y}\) and \(\Sigma_{y \rho}\), along with the covariance of \({\rho}\), are given by:
\begin{equation}
\Sigma_{\rho y} =
\begin{bmatrix}
k \mathcal{A}^* & k \mathcal{B}^*
\end{bmatrix},
\quad
\Sigma_{y \rho} = 
\begin{bmatrix}
\mathcal{A} k \\
\mathcal{B} k
\end{bmatrix},
\quad
\Sigma_{\rho \rho} = k.
\end{equation}
Applying the standard conditioning formula for multivariate Gaussian distributions, the posterior mean of the latent function 
\(\rho\) given the observations \(y\) can be expressed as:
\begin{equation}
\mathbb{E}[{\rho} | y] = 
\begin{bmatrix}
k \mathcal{A}^* & k \mathcal{B}^*
\end{bmatrix}
\begin{bmatrix}
\mathcal{A} k \mathcal{A}^* + \sigma^2 I & \mathcal{A} k \mathcal{B}^* \\
\mathcal{B} k \mathcal{A}^* & \mathcal{B} k \mathcal{B}^*
\end{bmatrix}^{-1}
y,
\end{equation}
and the posterior covariance is given by:
\begin{align}
\text{Cov}[{\rho} \mid y] 
= k & -
\begin{bmatrix}
k \mathcal{A}^* & k \mathcal{B}^*
\end{bmatrix} \nonumber \\
& \times
\begin{bmatrix}
\mathcal{A} k \mathcal{A}^* + \sigma^2 I & \mathcal{A} k \mathcal{B}^* \\
\mathcal{B} k \mathcal{A}^* & \mathcal{B} k \mathcal{B}^*
\end{bmatrix}^{-1}
\begin{bmatrix}
\mathcal{A} k \\
\mathcal{B} k
\end{bmatrix}.
\end{align}
The inverse of the joint covariance matrix \(\Sigma_{yy}\) can be obtained using the block matrix inversion formula:
\begin{equation}
\Sigma_{yy}^{-1} =
\begin{bmatrix}
P^{-1} + P^{-1} Q S_c^{-1} R P^{-1} & -P^{-1} Q S_c^{-1} \\
-S_c^{-1} R P^{-1} & S_c^{-1}
\end{bmatrix},
\end{equation}
where the submatrices are defined as: 
\[
P = \mathcal{A} k \mathcal{A}^* + \sigma^2 I, \quad Q = \mathcal{A} k \mathcal{B}^*, \quad R = \mathcal{B} k \mathcal{A}^*,\quad S = \mathcal{B} k \mathcal{B}^*,
\]
and the Schur complement is given by:
\[S_c = S - R P^{-1} Q.\]
The posterior mean, \(\mathbb{E}[{\rho} | y]\), is then obtained by conditioning on the observed data \(y\) using the block matrix inversion results. This leads to the following expression:
\begin{align}
\mathbb{E}[\rho \mid y] &= k \bigg( \mathcal{A}^* (\mathcal{A} k \mathcal{A}^* + \sigma^2 I)^{-1} \mu \notag\\
&\quad + \mathcal{B}^* \left(\mathcal{B} k \mathcal{B}^* - \mathcal{B} k \mathcal{A}^* (\mathcal{A} k \mathcal{A}^* + \sigma^2 I)^{-1} \mathcal{A} k \mathcal{B}^*\right)^{-1} \notag\\
&\quad \times \left(1 - \mathcal{B} k \mathcal{A}^* (\mathcal{A} k \mathcal{A}^* + \sigma^2 I)^{-1} \mu \right) \bigg).
\end{align}
Similarly, the posterior covariance, \(\text{Cov}[{\rho} | y]\), is given by:
\begin{align}
\text{Cov}[\rho \mid y] &= k - k \bigg( \mathcal{A}^* (\mathcal{A} k \mathcal{A}^* + \sigma^2 I)^{-1} \mathcal{A} k \notag\\
&\quad + \mathcal{B}^* \left(\mathcal{B} k \mathcal{B}^* - \mathcal{B} k \mathcal{A}^* (\mathcal{A} k \mathcal{A}^* + \sigma^2 I)^{-1} \mathcal{A} k \mathcal{B}^*\right)^{-1} \notag\\
&\quad \times \mathcal{B} k \bigg).
\end{align}
These expanded expressions clearly show the role of the measurement operator \(\mathcal{A}\), the constraint operator \(\mathcal{B}\), and the prior covariance \(k\). This structured formulation provides a clear interpretation of how the prior information and the physical constraints are integrated into the posterior distribution, ensuring that both the measurement model and normalization constraint are accurately incorporated into the final estimates.

\subsubsection{\label{sec:Ker_Ei_Ex}Posterior Estimation with Basis Function Approximation}
In this section, we extend the kernel eigenfunction expansion approach to incorporate constraints. By employing a low-rank approximation of the covariance kernel, this method maintains computational efficiency while explicitly incorporating the normalization constraint into the GP model. Similarly, the integral constraint can be enforced as:
\begin{equation}
1 = \mathcal{B} \Phi \alpha = \Psi_\mathcal{B} \alpha,
\end{equation}
where \(\Psi_\mathcal{B} \) represents the projection of the basis functions under the constraint operator, effectively mapping the eigenfunction expansion to the constraint domain. The elements of the joint covariance matrix \(\Sigma_{yy}\) in the basis function approximation in Eq.~\eqref{Sigma_yy} are given by:
\begin{equation}
\Sigma_{yy}
\approx
\begin{bmatrix}
\Psi_\mathcal{A} \Lambda \Psi_\mathcal{A}^\intercal + \sigma^2 I & \Psi_\mathcal{A} \Lambda \Psi_\mathcal{B}^\intercal \\
\Psi_\mathcal{B} \Lambda \Psi_\mathcal{A}^\intercal & \Psi_\mathcal{B} \Lambda \Psi_\mathcal{B}^\intercal
\end{bmatrix},
\label{eq:cova}
\end{equation}
where the cross-covariances and the prior covariance of \(\rho\) are given by:
\begin{equation}
\begin{aligned}
&\Sigma_{\rho y} 
\approx 
\begin{bmatrix}
\Phi \Lambda \Psi_\mathcal{A}^\intercal & \Phi \Lambda \Psi_\mathcal{B}^\intercal
\end{bmatrix},
\quad
\Sigma_{y \rho}
 \approx
\begin{bmatrix}
\Psi_\mathcal{A} \Lambda \Phi^\intercal \\
\Psi_\mathcal{B} \Lambda \Phi^\intercal
\end{bmatrix},
\quad \\
&\Sigma_{\rho\rho} \approx \Phi \Lambda \Phi^\intercal.
\end{aligned}
\end{equation}
The general conditional Gaussian formula gives the posterior mean and covariance of \(\rho\) given \(y\):
\begin{equation}
\begin{aligned}
\mathbb{E}[\rho \mid y]
= \; &
\begin{bmatrix}
\Phi \Lambda \Psi_\mathcal{A}^\intercal & \Phi \Lambda \Psi_\mathcal{B}^\intercal
\end{bmatrix} \\
& \times
\begin{bmatrix}
\Psi_\mathcal{A} \Lambda \Psi_\mathcal{A}^\intercal + \sigma^2 I & \Psi_\mathcal{A} \Lambda \Psi_\mathcal{B}^\intercal \\
\Psi_\mathcal{B} \Lambda \Psi_\mathcal{A}^\intercal & \Psi_\mathcal{B} \Lambda \Psi_\mathcal{B}^\intercal
\end{bmatrix}^{-1} 
\begin{bmatrix}
\mu \\
1
\end{bmatrix},
\end{aligned}
\end{equation}
\begin{equation}
\begin{aligned}
\text{Cov}[\rho \mid y]
= \; & \Phi \Lambda \Phi^\intercal \\
& - 
\begin{bmatrix}
\Phi \Lambda \Psi_\mathcal{A}^\intercal & \Phi \Lambda \Psi_\mathcal{B}^\intercal
\end{bmatrix} \\
& \times
\begin{bmatrix}
\Psi_\mathcal{A} \Lambda \Psi_\mathcal{A}^\intercal + \sigma^2 I & \Psi_\mathcal{A} \Lambda \Psi_\mathcal{B}^\intercal \\
\Psi_\mathcal{B} \Lambda \Psi_\mathcal{A}^\intercal & \Psi_\mathcal{B} \Lambda \Psi_\mathcal{B}^\intercal
\end{bmatrix}^{-1} \\
& \times
\begin{bmatrix}
\Psi_\mathcal{A} \Lambda \Phi^\intercal \\
\Psi_\mathcal{B} \Lambda \Phi^\intercal
\end{bmatrix}.
\end{aligned}
\end{equation}
By defining a combined projection matrix \(\Psi\) that captures both measurement and constraint operators in a unified representation:
\begin{equation}
\Psi = \begin{bmatrix} \Psi_\mathcal{A} \\ \Psi_\mathcal{B} \end{bmatrix},
\end{equation}
the posterior mean of the estimated distribution is given by:
\begin{equation}
\mathbb{E}[\rho \mid y] = \Phi \Lambda \Psi^\intercal \left(\Psi \Lambda \Psi^\intercal + N\right)^{-1} \begin{bmatrix} \mu \\ 1 \end{bmatrix}, N = \begin{bmatrix} \sigma^2 I & 0 \\ 0 & 0 \end{bmatrix},
\end{equation}
where \(N\) is a diagonal noise matrix that accounts for the measurement uncertainty in \(\mu\) and enforces the normalization constraint as noise-free. The posterior covariance is given by:
\begin{equation}
\text{Cov}[\rho \mid y] = \Phi \Lambda \Phi^\intercal - \Phi \Lambda \Psi^\intercal \left(\Psi \Lambda \Psi^\intercal +N\right)^{-1} \Psi \Lambda \Phi^\intercal.
\end{equation}

\subsection{\label{sec:Param_Est}Parameter Estimation}

To fully specify the GPR model, we must determine suitable values for the hyperparameters that define the covariance function and the noise model. In this section, we present two complementary strategies for hyperparameter estimation. Section~\ref{sec:Log_Mar} outlines a standard data-driven approach based on maximizing the unconstrained log marginal likelihood using the observed scattering data. In contrast, Section~\ref{sec:Cons_only} presents an approach in which physical constraints are
incorporated to estimate the kernel hyperparameters independently of the observed data.

\subsubsection{\label{sec:Log_Mar}Log Marginal Likelihood for Hyperparameter Optimization}

To optimize the hyperparameters of a GP model, we can use Bayesian methods based on marginal likelihood \cite{rasmussen2006gaussian}. Hyperparameters—such as the length scale, \(\ell\), kernel variance, \(\sigma_f\), in the SE covariance function, and the noise parameter, \(\sigma\)—are learned from the data while balancing model complexity,
\(
\theta = (\sigma_f, \ell, \sigma).
\)
Using Bayes’ theorem, the marginal likelihood of the data given hyperparameters \(\theta\) is computed by integrating out the unknown function \(\rho\):
\begin{equation}
p(\mu\mid\theta) = \int p (\mu \mid\rho, \theta)p(\rho\mid \theta) d\rho.
\end{equation}
The standard form of the marginal likelihood for a GP observing data \(\mu\) and hyperparameters \(\theta\) is expressed as:
\begin{equation}
p(\mu \mid \theta) = \mathcal{N} (\mu \mid 0, D),
\end{equation}
where 
\(
D = \Psi_\mathcal{A} \Lambda \Psi_\mathcal{A}^\intercal + \sigma^2 I.
\)
The posterior distribution over hyperparameters can now be written as follows: 
\begin{equation}
p(\theta \mid\mu) \propto p(\mu \mid \theta) \, p(\sigma_f)\, p(\ell)\, p(\sigma).
\end{equation}
The integration yields the logarithmic marginal likelihood;
\begin{equation}
\begin{aligned}
\log p(\mu \mid \theta) =& -\frac{1}{2} \log |D| - \frac{1}{2} \mu^\intercal D^{-1} \mu - \frac{n}{2} \log 2 \pi.
\end{aligned}
\end{equation}
This function is used to optimize the hyperparameters of the GP model. To determine the optimal hyperparameters \(\theta^*\), we maximize the log-marginal likelihood:
\begin{equation}
\theta^* = \arg\max_{\theta} \log p(\mu | \theta).
\end{equation}

\subsubsection{\label{sec:Cons_only}Joint Log Marginal Likelihood Incorporating Physical Constraints}

In the standard GPR framework, hyperparameters \(\theta = (\sigma_f, \ell, \sigma)\) are typically estimated by maximizing the marginal likelihood based solely on the observed scattering measurements \(\mu\). However, when the data are noisy, sparse, or incomplete, relying exclusively on the measurement model may result in unreliable or overfitted hyperparameter estimates. To enhance robustness and fully incorporate prior physical knowledge, we adopt a joint probabilistic model that integrates the observational data together with a known constraint on the solution: the normalization of the size distribution \(\rho(r)\), expressed as \(Z = 1\).

The joint distribution of the measurement data \(\mu\) and the constraint \(Z\) is modeled as a multivariate Gaussian:
\[y=
\begin{bmatrix}
\mu \\ Z
\end{bmatrix}
\sim \mathcal{N}\left(
\begin{bmatrix}
0 \\ 0
\end{bmatrix},
\Sigma_{yy}
\right),
\]
where the combined covariance matrix \(\Sigma_{yy}\) is defined as:

\begin{equation}
\Sigma_{yy}=
\begin{bmatrix}
\Psi_\mathcal{A} \Lambda \Psi_\mathcal{A}^\top + \sigma^2 I & \Psi_\mathcal{A} \Lambda \Psi_\mathcal{B}^\top \\
\Psi_\mathcal{B} \Lambda \Psi_\mathcal{A}^\top & \Psi_\mathcal{B} \Lambda \Psi_\mathcal{B}^\top
\end{bmatrix}.
\end{equation}
The joint log marginal likelihood of the measurements and constraint is expressed as:
\begin{equation}
\log p(y \mid \theta) = -\frac{1}{2} \log |\Sigma_{yy}| - \frac{1}{2} y^\top \Sigma_{yy}^{-1} y - \frac{n+1}{2} \log (2\pi),
\end{equation}
where \(n\) is the number of measurement points in \(\mu\), and the total dimension of \(y\) is \(n+1\) due to the inclusion of the constraint.

To facilitate interpretation and efficient computation, we define the following matrix blocks:
\begin{align}
D &= \Psi_\mathcal{A} \Lambda \Psi_\mathcal{A}^\top + \sigma^2 I, \\
V &= \Psi_\mathcal{A} \Lambda \Psi_\mathcal{B}^\top, \\
Y &= \Psi_\mathcal{B} \Lambda \Psi_\mathcal{B}^\top 
- \Psi_\mathcal{B} \Lambda \Psi_\mathcal{A}^\top D^{-1} \Psi_\mathcal{A} \Lambda \Psi_\mathcal{B}^\top.
\end{align}
Using the Schur complement and completing the square, the joint log marginal likelihood can be equivalently written as:
\begin{equation}
\begin{aligned}
\log p(y \mid \theta)
&= -\frac{1}{2} \mu^\top D^{-1} \mu \\
&\quad - \frac{1}{2} \left( 1 - V^\top D^{-1} \mu \right)^\top Y^{-1}\left( 1 - V^\top D^{-1} \mu \right) \\
&\quad - \frac{1}{2} \log |D| - \frac{1}{2} \log |Y| \\
&\quad - \frac{n + 1}{2} \log (2\pi).
\end{aligned}
\end{equation}

This formulation unifies observational data and physical constraints within a principled Bayesian framework. Maximizing the joint log marginal likelihood leads to hyperparameter estimates \(\theta\) that are simultaneously informed by the measurements and the normalization condition.

\section{\label{sec:Result}Results and discussion} 

In this section, we present the results of the constrained GPR framework applied to the estimation of the particle size distribution, $\rho(r)$, from scattering coefficients, $\mu_{\text{sca}}(\lambda)$. The analysis systematically evaluates the impact of enforcing physical constraints, the role of hyperparameter selection, the effect of different covariance kernels, and the correspondence between measured and predicted scattering coefficients.

First, we assess the importance of imposing the normalization constraint on the estimated $\rho(r)$ to ensure physically meaningful solutions. Then, we examine the performance of the model in hyperparameter estimation using the joint log marginal likelihood approach. The analysis includes the application of two covariance kernels within the GPR methodology. Finally, we compare the measured and computed scattering coefficients, $\mu(\lambda)$, to verify the accuracy and consistency of the estimated particle size distributions.

This study utilizes a hybrid data approach to evaluate the proposed constrained GPR approach. The true particle size distribution, $\rho(r)$, is derived from experimental static light scattering (SLS) measurements~\cite{conley2021silica}, while the scattering data, $\mu_{\text{sca}}(\lambda)$, is synthetically generated using Lorenz-Mie theory due to the unavailability of direct measurements. Consequently, the dataset is semi-experimental, combining real and simulated components to enable evaluation of the inverse method.

\subsection{\label{sec:Impa}Impact of Constraint on the Estimated $\rho(r)$}

In our inverse problem, it is crucial to enforce the normalization constraint that $\rho(r)$ integrates to unity, in order to obtain physically meaningful solutions.

\begin{figure}[htbp]
\centering
\includegraphics[width=0.45\textwidth]{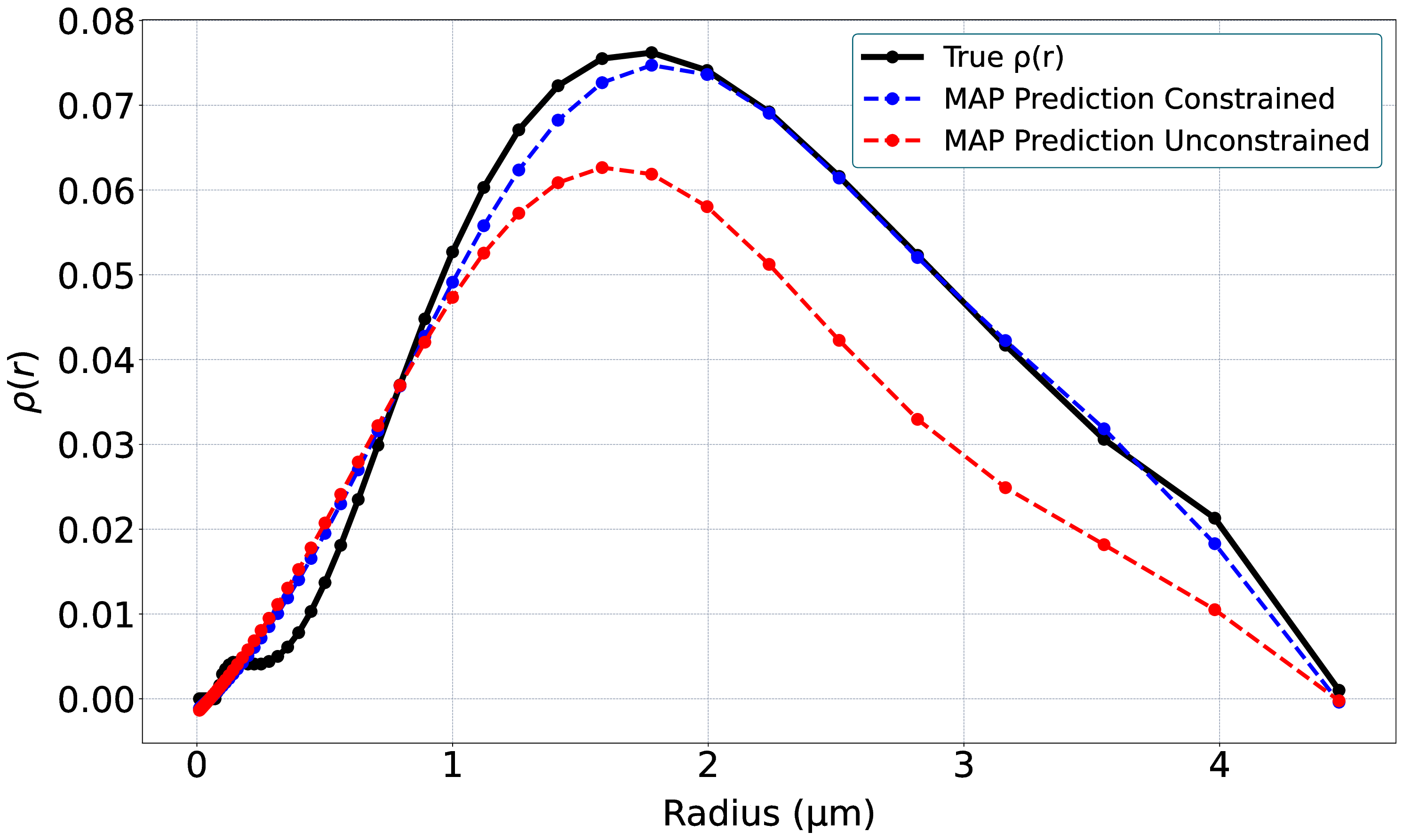}
\caption{Comparison of the estimated particle size distribution $\rho(r)$ based on the MAP solution with and without enforcing the constraint. The true distribution (black solid line) provides the reference, while the constrained (blue dashed line) and unconstrained (red dashed line) predictions illustrate the impact of enforcing physical constraints in the inversion process.}
\label{fig:Constraint}
\end{figure}

Figure~\ref{fig:Constraint} illustrates the significant impact of this constraint on size distribution estimation using GPR. The true distribution (black line) is compared to two estimated distributions: one from the constrained GPR model (blue dashed line), which incorporates the normalization directly via a Lagrange multiplier, and another from an unconstrained model (red dashed line). The constrained solution closely matches the true distribution, accurately capturing both amplitude and shape throughout the domain ($\sum \rho = 1.00$, MSE $= 6.41 \times 10^{-6}$, where MSE denotes the mean squared error).

Conversely, the unconstrained solution exhibits noticeable deviations, particularly in amplitude, where it systematically underestimates the expected behavior. Furthermore, without applying the constraint, the integral of $\rho(r)$ deviates from unity ($\sum \rho = 0.87$, MSE $= 4.80 \times 10^{-5}$).

These results confirm that incorporating the normalization constraint effectively stabilizes the inversion process, ensuring physical consistency of the solution and significantly improving the accuracy and interpretability of the estimated size distribution.

\subsection{\label{sec:Comparson}Hyperparameter Estimation}
 
\begin{figure}[h]
\centering
\includegraphics[width=0.45\textwidth]{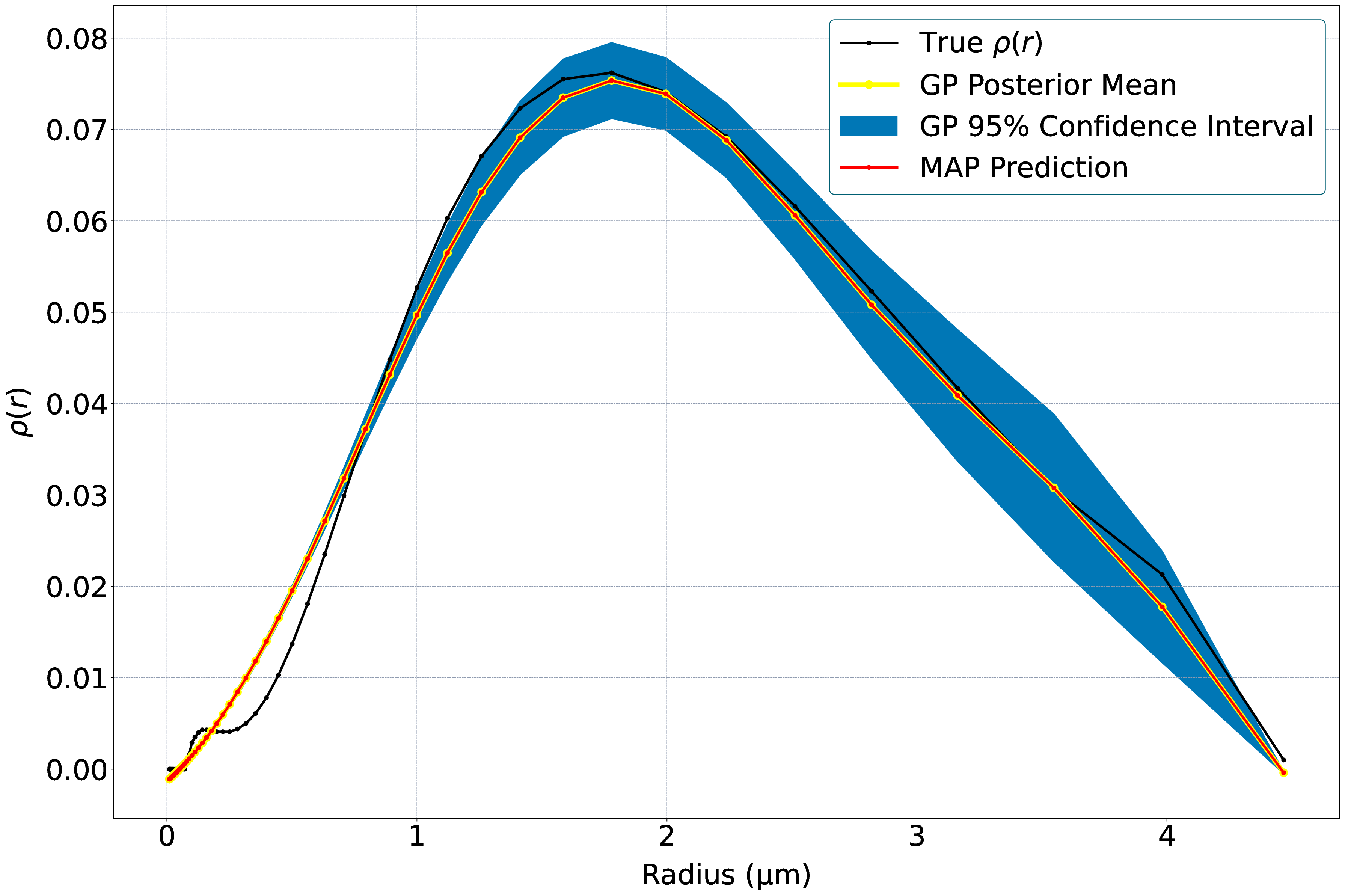}
\caption{Posterior inference of the particle size distribution estimated using the GP inverse formulation (yellow line). The shaded light blue region illustrates the 95\% confidence interval associated with the GP estimation. The red line indicates the Maximum a Posteriori (MAP) estimator. The true distribution (black solid line) provides the reference.}
\label{fig:rho_estimation_result}
\end{figure}

We estimate the particle size distribution \(\rho(r)\) using a constrained GPR model that incorporates both scattering measurements and a normalization constraint. This approach yields a smooth posterior mean estimate of \(\rho(r)\) along with quantified uncertainty from the posterior covariance. The GP hyperparameters are determined by maximizing the joint log marginal likelihood. Figure~\ref{fig:rho_estimation_result} illustrates the estimated distribution, where the yellow line denotes the posterior mean and the shaded region indicates the 95\% confidence interval.

A Maximum a Posteriori (MAP) solution is also computed by explicitly enforcing the normalization constraint using Lagrange multipliers in the optimization. The resulting MAP estimate (red line) aligns closely with the GP posterior mean, suggesting that the learned GP prior inherently encodes the correct physical structure of the problem. This consistency highlights the effectiveness of the joint log marginal likelihood framework in hyperparameter selection and posterior inference.

\subsection{\label{sec:Re_size}Comparison of Covariance Kernels in Size Distribution Estimation}

\begin{figure}[htbp]
\centering
\includegraphics[width=0.45\textwidth]{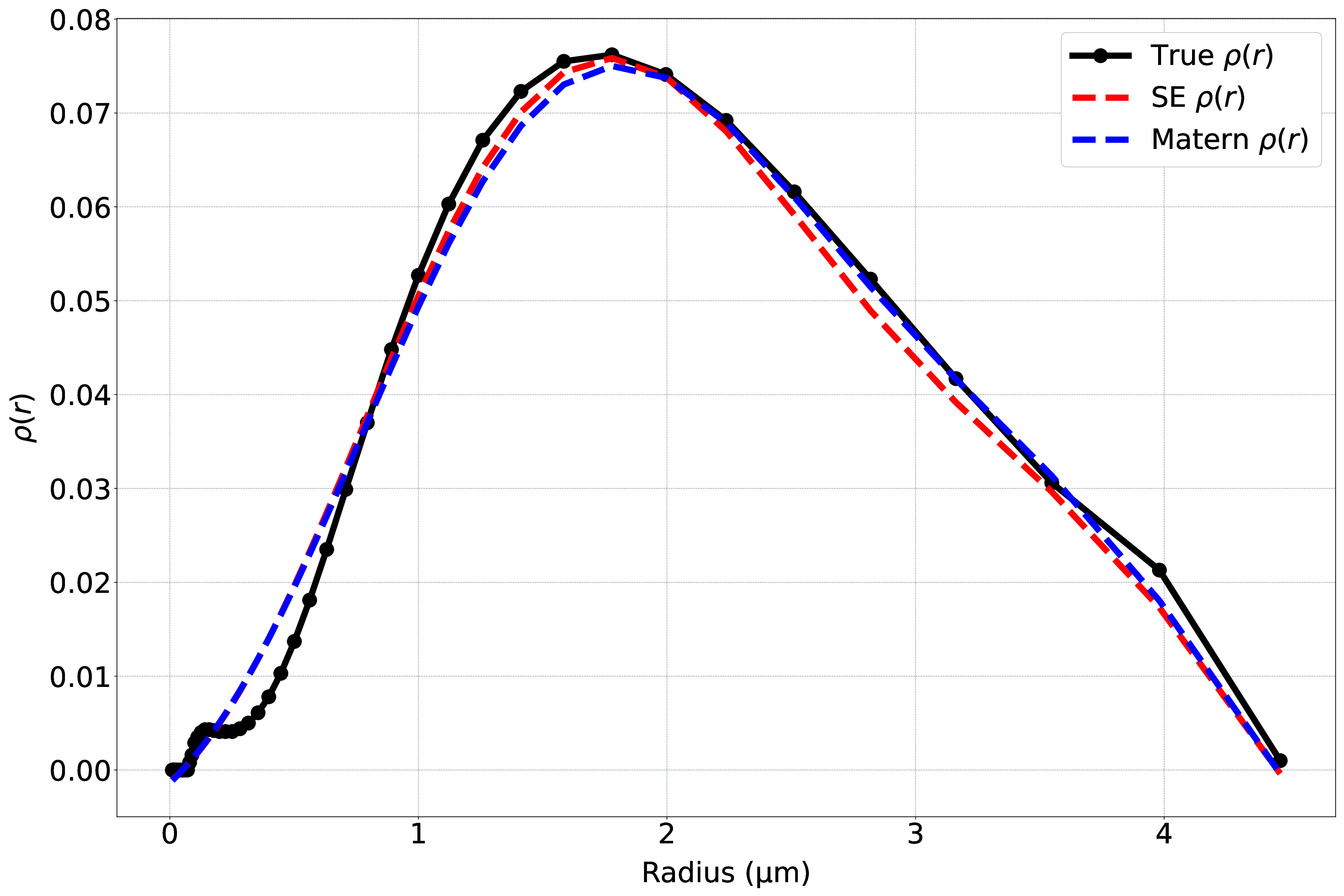}
\caption{Comparison of estimated solutions obtained using SE (red dashed line) and Matérn (blue dashed line) kernel functions. The true distribution (black solid line) provides the reference.}
\label{fig:Matern}
\end{figure}

To assess the influence of the covariance kernel on the reconstruction of particle size distributions, we compare the performance of two GP priors based on the SE and Mat\'ern kernels. Figure~\ref{fig:Matern} shows the estimated particle size distributions, $\rho(r)$, obtained using each kernel. In both cases, the estimated distributions closely match the true distribution, effectively capturing its overall shape, peak location, and spread. The similarity between the SE and Mat\'ern estimates and the true distribution demonstrates the robustness of the GP-based inverse formulation in recovering $\rho(r)$.
To quantitatively evaluate the estimation accuracy, we compute the mean squared error (MSE) between the estimated and true distributions. The MSE for the SE-based reconstruction is $6.29 \times 10^{-6}$, while for the Mat\'ern kernel it is $6.41 \times 10^{-6}$, indicating comparable performance. This comparison highlights the flexibility of the GP approach in incorporating different prior assumptions through the kernel function and confirms that both kernels produce accurate and physically consistent solutions.

\subsection{\label{sec:Comparison}Comparison of Observed and Computed Scattering Coefficients}

\begin{figure}[htbp]
\centering
\includegraphics[width=0.45\textwidth]{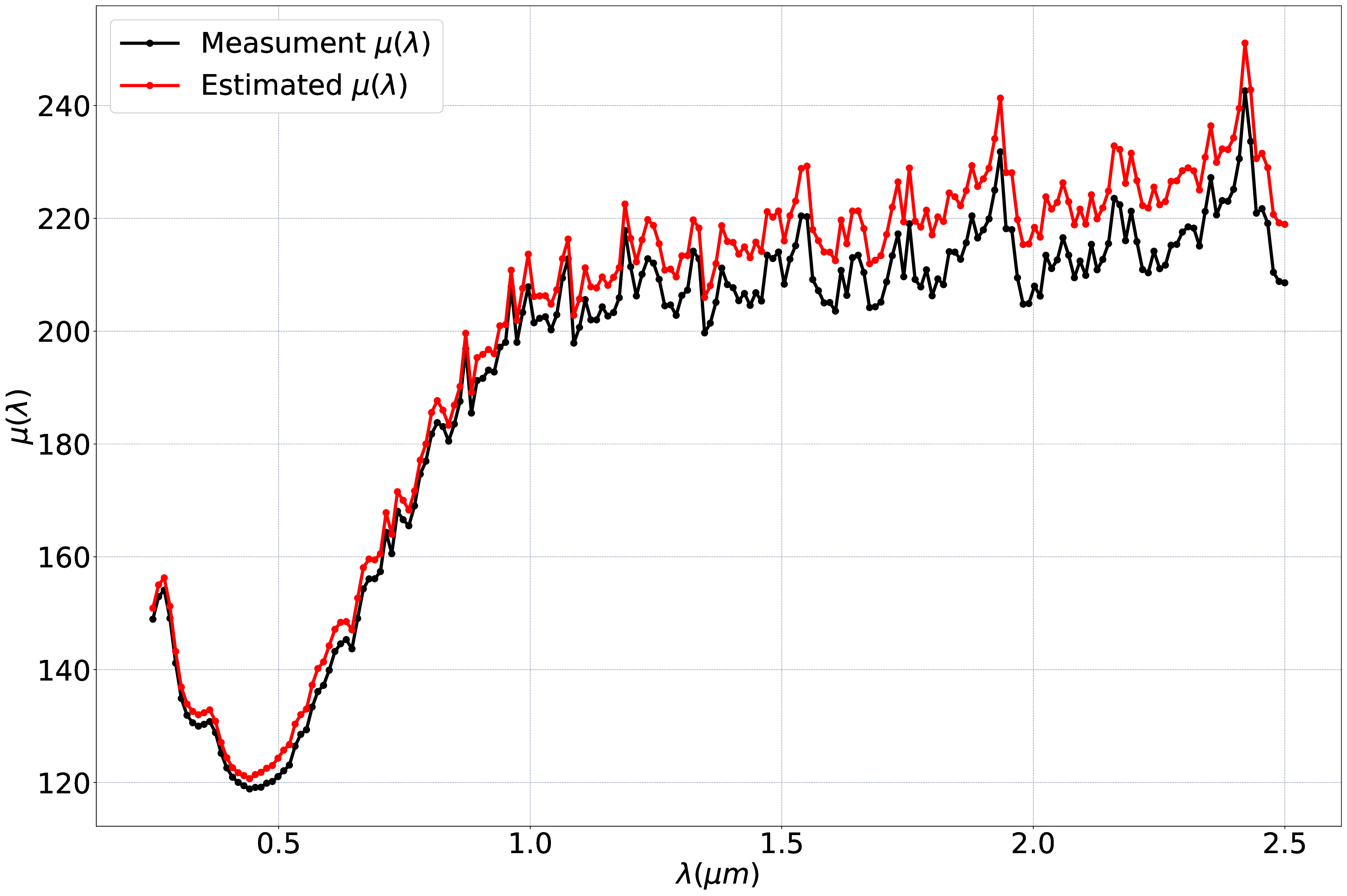}
\caption{Comparison of measured (black line) and estimated (red line) scattering coefficients, $\mu(\lambda)$.}
\label{fig:Comparison}
\end{figure}

As part of this study, we validate the estimated particle size distribution, $\rho(r)$, by evaluating its ability to reproduce the experimentally measured scattering coefficients, $\mu(\lambda)$. This is achieved by substituting the estimated distribution into the forward model defined by the integral equation and comparing the predicted values of $\mu(\lambda)$ with the measured data. Figure~\ref{fig:Comparison} illustrates this comparison and confirms that it successfully reproduces the experimental observations.
The results remain consistent across the entire wavelength range, showing that the inverse method produces reliable predictions. This consistency suggests that the estimated distribution and the modeling framework work well together to describe the scattering behavior observed in the measurements.

\section{\label{sec:Conclu}Conclusion}

This study demonstrates the effectiveness of the constrained GPR framework for estimating particle size distributions (PSDs) from optical scattering data. By modeling the inverse problem as a Fredholm integral equation of the first kind, we successfully reconstructed smooth and physically meaningful PSDs. The incorporation of constraints via Lagrange multipliers ensures that essential physical requirements, such as normalization, are strictly enforced, resulting in improved solution stability and interpretability.

The combination of the GPR framework with regularized least squares allows for flexible and robust estimation of PSDs, even in the presence of noisy or sparse data. By leveraging the covariance structure of the GP prior, the method naturally incorporates smoothness assumptions and provides uncertainty quantification through posterior confidence intervals. Additionally, the application of the Woodbury matrix identity enhances computational efficiency, enabling efficient inversion of large covariance matrices that arise in high-dimensional problems.

A key strength of the proposed approach is its ability to meet the Hadamard conditions of well-posedness: existence, uniqueness, and stability of the solution. By enforcing physically consistent constraints and incorporating measurement uncertainties, the method produces reliable and interpretable estimates that are robust against small perturbations in the input data. This capability is particularly important for practical applications in particle characterization, where experimental data may often be limited or noisy.

Future work could extend this methodology by exploring alternative GP kernel functions or hierarchical Bayesian frameworks, which may further enhance the model’s adaptability to a broader range of particle systems and scattering regimes. Additionally, applying the method to purely experimental datasets—beyond the semi-synthetic data used in this study—would demonstrate its robustness and applicability under real measurement conditions. Such integration would strengthen its role in practical particle characterization workflows, facilitating seamless adoption in experimental laboratories and industrial environments.

\begin{acknowledgments}
The AI-TranspWood project, HORIZON-CL4–2023-RESILIENCE- 01–23 (Grant Agreement 101138191), is gratefully acknowledged. This project is co-funded by the European Union. Views and opinions expressed are however those of the author(s) only and do not necessarily reflect those of the European Union or HaDEA. Neither the European Union nor the granting authority can be held responsible for them.
\end{acknowledgments}

\bibliography{main}

\begin{thebibliography}{40}%
\makeatletter
\providecommand \@ifxundefined [1]{%
 \@ifx{#1\undefined}
}%
\providecommand \@ifnum [1]{%
 \ifnum #1\expandafter \@firstoftwo
 \else \expandafter \@secondoftwo
 \fi
}%
\providecommand \@ifx [1]{%
 \ifx #1\expandafter \@firstoftwo
 \else \expandafter \@secondoftwo
 \fi
}%
\providecommand \natexlab [1]{#1}%
\providecommand \enquote  [1]{``#1''}%
\providecommand \bibnamefont  [1]{#1}%
\providecommand \bibfnamefont [1]{#1}%
\providecommand \citenamefont [1]{#1}%
\providecommand \href@noop [0]{\@secondoftwo}%
\providecommand \href [0]{\begingroup \@sanitize@url \@href}%
\providecommand \@href[1]{\@@startlink{#1}\@@href}%
\providecommand \@@href[1]{\endgroup#1\@@endlink}%
\providecommand \@sanitize@url [0]{\catcode `\\12\catcode `\$12\catcode `\&12\catcode `\#12\catcode `\^12\catcode `\_12\catcode `\%12\relax}%
\providecommand \@@startlink[1]{}%
\providecommand \@@endlink[0]{}%
\providecommand \url  [0]{\begingroup\@sanitize@url \@url }%
\providecommand \@url [1]{\endgroup\@href {#1}{\urlprefix }}%
\providecommand \urlprefix  [0]{URL }%
\providecommand \Eprint [0]{\href }%
\providecommand \doibase [0]{https://doi.org/}%
\providecommand \selectlanguage [0]{\@gobble}%
\providecommand \bibinfo  [0]{\@secondoftwo}%
\providecommand \bibfield  [0]{\@secondoftwo}%
\providecommand \translation [1]{[#1]}%
\providecommand \BibitemOpen [0]{}%
\providecommand \bibitemStop [0]{}%
\providecommand \bibitemNoStop [0]{.\EOS\space}%
\providecommand \EOS [0]{\spacefactor3000\relax}%
\providecommand \BibitemShut  [1]{\csname bibitem#1\endcsname}%
\let\auto@bib@innerbib\@empty
%</preamble>
\bibitem [{\citenamefont {Allen}(2013)}]{allen2013particle}%
  \BibitemOpen
  \bibfield  {author} {\bibinfo {author} {\bibfnamefont {T.}~\bibnamefont {Allen}},\ }\href@noop {} {\emph {\bibinfo {title} {Particle size measurement}}}\ (\bibinfo  {publisher} {Springer},\ \bibinfo {year} {2013})\BibitemShut {NoStop}%
\bibitem [{\citenamefont {Willeke}\ and\ \citenamefont {Whitby}(1983)}]{willeke1975atmospheric}%
  \BibitemOpen
  \bibfield  {author} {\bibinfo {author} {\bibfnamefont {K.}~\bibnamefont {Willeke}}\ and\ \bibinfo {author} {\bibfnamefont {K.~T.}\ \bibnamefont {Whitby}},\ }\bibfield  {title} {\bibinfo {title} {Atmospheric aerosols: size distribution interpretation},\ }\href@noop {} {\bibfield  {journal} {\bibinfo  {journal} {J. Air Pollut. Control Assoc.}\ }\textbf {\bibinfo {volume} {25}},\ \bibinfo {pages} {529} (\bibinfo {year} {1983})}\BibitemShut {NoStop}%
\bibitem [{\citenamefont {Shekunov}\ \emph {et~al.}(2007)\citenamefont {Shekunov}, \citenamefont {Chattopadhyay}, \citenamefont {Tong},\ and\ \citenamefont {Chow}}]{shekunov2007particle}%
  \BibitemOpen
  \bibfield  {author} {\bibinfo {author} {\bibfnamefont {B.~Y.}\ \bibnamefont {Shekunov}}, \bibinfo {author} {\bibfnamefont {P.}~\bibnamefont {Chattopadhyay}}, \bibinfo {author} {\bibfnamefont {H.~H.~Y.}\ \bibnamefont {Tong}},\ and\ \bibinfo {author} {\bibfnamefont {A.~H.~L.}\ \bibnamefont {Chow}},\ }\bibfield  {title} {\bibinfo {title} {Particle size analysis in pharmaceutics: principles, methods and applications},\ }\href@noop {} {\bibfield  {journal} {\bibinfo  {journal} {Pharm. Res.}\ }\textbf {\bibinfo {volume} {24}},\ \bibinfo {pages} {203} (\bibinfo {year} {2007})}\BibitemShut {NoStop}%
\bibitem [{\citenamefont {Gupta}\ and\ \citenamefont {Sathiyamoorthy}(1998)}]{gupta1998fluid}%
  \BibitemOpen
  \bibfield  {author} {\bibinfo {author} {\bibfnamefont {C.~K.}\ \bibnamefont {Gupta}}\ and\ \bibinfo {author} {\bibfnamefont {D.}~\bibnamefont {Sathiyamoorthy}},\ }\href@noop {} {\emph {\bibinfo {title} {Fluid bed technology in materials processing}}}\ (\bibinfo  {publisher} {CRC press},\ \bibinfo {year} {1998})\BibitemShut {NoStop}%
\bibitem [{\citenamefont {Bohren}\ and\ \citenamefont {Huffman}(2008)}]{bohren2008absorption}%
  \BibitemOpen
  \bibfield  {author} {\bibinfo {author} {\bibfnamefont {C.~F.}\ \bibnamefont {Bohren}}\ and\ \bibinfo {author} {\bibfnamefont {D.~R.}\ \bibnamefont {Huffman}},\ }\href@noop {} {\emph {\bibinfo {title} {Absorption and scattering of light by small particles}}}\ (\bibinfo  {publisher} {John Wiley \& Sons},\ \bibinfo {year} {2008})\BibitemShut {NoStop}%
\bibitem [{\citenamefont {Mishchenko}\ \emph {et~al.}(2002)\citenamefont {Mishchenko}, \citenamefont {Travis},\ and\ \citenamefont {Lacis}}]{mishchenko2002scattering}%
  \BibitemOpen
  \bibfield  {author} {\bibinfo {author} {\bibfnamefont {M.~I.}\ \bibnamefont {Mishchenko}}, \bibinfo {author} {\bibfnamefont {L.~D.}\ \bibnamefont {Travis}},\ and\ \bibinfo {author} {\bibfnamefont {A.~A.}\ \bibnamefont {Lacis}},\ }\href@noop {} {\emph {\bibinfo {title} {Scattering, absorption, and emission of light by small particles}}}\ (\bibinfo  {publisher} {Cambridge university press},\ \bibinfo {year} {2002})\BibitemShut {NoStop}%
\bibitem [{\citenamefont {van~de Hulst}(1981)}]{hulst1981light}%
  \BibitemOpen
  \bibfield  {author} {\bibinfo {author} {\bibfnamefont {H.~C.}\ \bibnamefont {van~de Hulst}},\ }\href@noop {} {\emph {\bibinfo {title} {Light scattering by small particles}}}\ (\bibinfo  {publisher} {Courier Corporation},\ \bibinfo {year} {1981})\BibitemShut {NoStop}%
\bibitem [{\citenamefont {Twomey}(2019)}]{twomey2019introduction}%
  \BibitemOpen
  \bibfield  {author} {\bibinfo {author} {\bibfnamefont {S.}~\bibnamefont {Twomey}},\ }\href@noop {} {\emph {\bibinfo {title} {Introduction to the mathematics of inversion in remote sensing and indirect measurements}}}\ (\bibinfo  {publisher} {Courier Dover Publications},\ \bibinfo {year} {2019})\BibitemShut {NoStop}%
\bibitem [{\citenamefont {Colton}\ and\ \citenamefont {Kress}(1998)}]{colton1998inverse}%
  \BibitemOpen
  \bibfield  {author} {\bibinfo {author} {\bibfnamefont {D.~L.}\ \bibnamefont {Colton}}\ and\ \bibinfo {author} {\bibfnamefont {R.}~\bibnamefont {Kress}},\ }\href@noop {} {\emph {\bibinfo {title} {Inverse acoustic and electromagnetic scattering theory}}},\ Vol.~\bibinfo {volume} {93}\ (\bibinfo  {publisher} {Springer},\ \bibinfo {year} {1998})\BibitemShut {NoStop}%
\bibitem [{\citenamefont {Holt}\ \emph {et~al.}(1978)\citenamefont {Holt}, \citenamefont {Uzunoglu},\ and\ \citenamefont {Evans}}]{holt1978integral}%
  \BibitemOpen
  \bibfield  {author} {\bibinfo {author} {\bibfnamefont {A.}~\bibnamefont {Holt}}, \bibinfo {author} {\bibfnamefont {N.}~\bibnamefont {Uzunoglu}},\ and\ \bibinfo {author} {\bibfnamefont {B.}~\bibnamefont {Evans}},\ }\bibfield  {title} {\bibinfo {title} {An integral equation solution to the scattering of electromagnetic radiation by dielectric spheroids and ellipsoids},\ }\href@noop {} {\bibfield  {journal} {\bibinfo  {journal} {IEEE Trans. Antennas Propag.}\ }\textbf {\bibinfo {volume} {26}},\ \bibinfo {pages} {706} (\bibinfo {year} {1978})}\BibitemShut {NoStop}%
\bibitem [{\citenamefont {Twomey}\ and\ \citenamefont {Severynse}(1963)}]{twomey1963measurements}%
  \BibitemOpen
  \bibfield  {author} {\bibinfo {author} {\bibfnamefont {S.}~\bibnamefont {Twomey}}\ and\ \bibinfo {author} {\bibfnamefont {G.~T.}\ \bibnamefont {Severynse}},\ }\bibfield  {title} {\bibinfo {title} {Measurements of size distributions of natural aerosols},\ }\href@noop {} {\bibfield  {journal} {\bibinfo  {journal} {J. Atmos. Sci.}\ }\textbf {\bibinfo {volume} {20}},\ \bibinfo {pages} {392} (\bibinfo {year} {1963})}\BibitemShut {NoStop}%
\bibitem [{\citenamefont {Yamamoto}\ and\ \citenamefont {Tanaka}(1969)}]{yamamoto1969determination}%
  \BibitemOpen
  \bibfield  {author} {\bibinfo {author} {\bibfnamefont {G.}~\bibnamefont {Yamamoto}}\ and\ \bibinfo {author} {\bibfnamefont {M.}~\bibnamefont {Tanaka}},\ }\bibfield  {title} {\bibinfo {title} {Determination of aerosol size distribution from spectral attenuation measurements},\ }\href@noop {} {\bibfield  {journal} {\bibinfo  {journal} {Appl. Opt.}\ }\textbf {\bibinfo {volume} {8}},\ \bibinfo {pages} {447} (\bibinfo {year} {1969})}\BibitemShut {NoStop}%
\bibitem [{\citenamefont {John}(1991)}]{john1991partial}%
  \BibitemOpen
  \bibfield  {author} {\bibinfo {author} {\bibfnamefont {F.}~\bibnamefont {John}},\ }\href@noop {} {\emph {\bibinfo {title} {Partial differential equations}}},\ Vol.~\bibinfo {volume} {1}\ (\bibinfo  {publisher} {Springer Science \& Business Media},\ \bibinfo {year} {1991})\BibitemShut {NoStop}%
\bibitem [{\citenamefont {Tikhonov}\ and\ \citenamefont {Arsenin}(1977)}]{tikhonov1977solutions}%
  \BibitemOpen
  \bibfield  {author} {\bibinfo {author} {\bibfnamefont {A.~N.}\ \bibnamefont {Tikhonov}}\ and\ \bibinfo {author} {\bibfnamefont {V.}~\bibnamefont {Arsenin}},\ }\href@noop {} {\emph {\bibinfo {title} {Solutions of ill-posed problems}}}\ (\bibinfo  {publisher} {V. H. Winston},\ \bibinfo {year} {1977})\BibitemShut {NoStop}%
\bibitem [{\citenamefont {Phillips}(1962)}]{phillips1962technique}%
  \BibitemOpen
  \bibfield  {author} {\bibinfo {author} {\bibfnamefont {D.~L.}\ \bibnamefont {Phillips}},\ }\bibfield  {title} {\bibinfo {title} {A technique for the numerical solution of certain integral equations of the first kind},\ }\href@noop {} {\bibfield  {journal} {\bibinfo  {journal} {J. ACM}\ }\textbf {\bibinfo {volume} {9}},\ \bibinfo {pages} {84} (\bibinfo {year} {1962})}\BibitemShut {NoStop}%
\bibitem [{\citenamefont {Tikhonov}(1963)}]{tikhonov1963solution}%
  \BibitemOpen
  \bibfield  {author} {\bibinfo {author} {\bibfnamefont {A.~N.}\ \bibnamefont {Tikhonov}},\ }\bibfield  {title} {\bibinfo {title} {Solution of incorrectly formulated problems and the regularization method},\ }\href@noop {} {\bibfield  {journal} {\bibinfo  {journal} {Sov. Math. Dokl.}\ }\textbf {\bibinfo {volume} {4}},\ \bibinfo {pages} {1035} (\bibinfo {year} {1963})}\BibitemShut {NoStop}%
\bibitem [{\citenamefont {Engl}\ \emph {et~al.}(1996)\citenamefont {Engl}, \citenamefont {Hanke},\ and\ \citenamefont {Neubauer}}]{engl1996regularization}%
  \BibitemOpen
  \bibfield  {author} {\bibinfo {author} {\bibfnamefont {H.~W.}\ \bibnamefont {Engl}}, \bibinfo {author} {\bibfnamefont {M.}~\bibnamefont {Hanke}},\ and\ \bibinfo {author} {\bibfnamefont {A.}~\bibnamefont {Neubauer}},\ }\href@noop {} {\emph {\bibinfo {title} {Regularization of inverse problems}}},\ Vol.\ \bibinfo {volume} {375}\ (\bibinfo  {publisher} {Springer Science \& Business Media},\ \bibinfo {year} {1996})\BibitemShut {NoStop}%
\bibitem [{\citenamefont {Rasmussen}\ and\ \citenamefont {Williams}(2006)}]{rasmussen2006gaussian}%
  \BibitemOpen
  \bibfield  {author} {\bibinfo {author} {\bibfnamefont {C.~E.}\ \bibnamefont {Rasmussen}}\ and\ \bibinfo {author} {\bibfnamefont {C.~K.~I.}\ \bibnamefont {Williams}},\ }\href@noop {} {\emph {\bibinfo {title} {Gaussian processes for machine learning}}}\ (\bibinfo  {publisher} {MIT press},\ \bibinfo {year} {2006})\BibitemShut {NoStop}%
\bibitem [{\citenamefont {Bishop}\ and\ \citenamefont {Nasrabadi}(2006)}]{bishop2006pattern}%
  \BibitemOpen
  \bibfield  {author} {\bibinfo {author} {\bibfnamefont {C.~M.}\ \bibnamefont {Bishop}}\ and\ \bibinfo {author} {\bibfnamefont {N.~M.}\ \bibnamefont {Nasrabadi}},\ }\href@noop {} {\emph {\bibinfo {title} {Pattern recognition and machine learning}}}\ (\bibinfo  {publisher} {Springer},\ \bibinfo {year} {2006})\BibitemShut {NoStop}%
\bibitem [{\citenamefont {Swiler}\ \emph {et~al.}(2020)\citenamefont {Swiler}, \citenamefont {Gulian}, \citenamefont {Frankel}, \citenamefont {Safta},\ and\ \citenamefont {Jakeman}}]{swiler2020survey}%
  \BibitemOpen
  \bibfield  {author} {\bibinfo {author} {\bibfnamefont {L.~P.}\ \bibnamefont {Swiler}}, \bibinfo {author} {\bibfnamefont {M.}~\bibnamefont {Gulian}}, \bibinfo {author} {\bibfnamefont {A.~L.}\ \bibnamefont {Frankel}}, \bibinfo {author} {\bibfnamefont {C.}~\bibnamefont {Safta}},\ and\ \bibinfo {author} {\bibfnamefont {J.~D.}\ \bibnamefont {Jakeman}},\ }\bibfield  {title} {\bibinfo {title} {A survey of constrained {G}aussian process regression: Approaches and implementation challenges},\ }\href@noop {} {\bibfield  {journal} {\bibinfo  {journal} {J. Mach. Learn. Model. Comput.}\ }\textbf {\bibinfo {volume} {1}},\ \bibinfo {pages} {119} (\bibinfo {year} {2020})}\BibitemShut {NoStop}%
\bibitem [{\citenamefont {Solin}\ and\ \citenamefont {Kok}(2019)}]{solin2019know}%
  \BibitemOpen
  \bibfield  {author} {\bibinfo {author} {\bibfnamefont {A.}~\bibnamefont {Solin}}\ and\ \bibinfo {author} {\bibfnamefont {M.}~\bibnamefont {Kok}},\ }\bibfield  {title} {\bibinfo {title} {Know your boundaries: Constraining {G}aussian processes by variational harmonic features},\ }in\ \href@noop {} {\emph {\bibinfo {booktitle} {The 22nd International Conference on Artificial Intelligence and Statistics}}},\ Vol.~\bibinfo {volume} {89},\ \bibinfo {editor} {edited by\ \bibinfo {editor} {\bibfnamefont {K.}~\bibnamefont {Chaudhuri}}\ and\ \bibinfo {editor} {\bibfnamefont {M.}~\bibnamefont {Sugiyama}}}\ (\bibinfo  {publisher} {Proceedings of Machine Learning Research},\ \bibinfo {year} {2019})\ pp.\ \bibinfo {pages} {2193--2202}\BibitemShut {NoStop}%
\bibitem [{\citenamefont {Solin}\ and\ \citenamefont {S{\"a}rkk{\"a}}(2020)}]{solin2020hilbert}%
  \BibitemOpen
  \bibfield  {author} {\bibinfo {author} {\bibfnamefont {A.}~\bibnamefont {Solin}}\ and\ \bibinfo {author} {\bibfnamefont {S.}~\bibnamefont {S{\"a}rkk{\"a}}},\ }\bibfield  {title} {\bibinfo {title} {Hilbert space methods for reduced-rank {G}aussian process regression},\ }\href@noop {} {\bibfield  {journal} {\bibinfo  {journal} {Stat. Comput.}\ }\textbf {\bibinfo {volume} {30}},\ \bibinfo {pages} {419} (\bibinfo {year} {2020})}\BibitemShut {NoStop}%
\bibitem [{\citenamefont {Graepel}(2003)}]{graepel2003solving}%
  \BibitemOpen
  \bibfield  {author} {\bibinfo {author} {\bibfnamefont {T.}~\bibnamefont {Graepel}},\ }\bibfield  {title} {\bibinfo {title} {Solving noisy linear operator equations by {G}aussian processes: Application to ordinary and partial differential equations},\ }in\ \href@noop {} {\emph {\bibinfo {booktitle} {The 20th International Conference on International Conference on Machine Learning}}},\ \bibinfo {series} {ICML}, Vol.~\bibinfo {volume} {3}\ (\bibinfo  {publisher} {AAAI Press},\ \bibinfo {year} {2003})\ pp.\ \bibinfo {pages} {234--241}\BibitemShut {NoStop}%
\bibitem [{\citenamefont {Kelly}\ and\ \citenamefont {Rice}(1990)}]{kelly1990monotone}%
  \BibitemOpen
  \bibfield  {author} {\bibinfo {author} {\bibfnamefont {C.}~\bibnamefont {Kelly}}\ and\ \bibinfo {author} {\bibfnamefont {J.}~\bibnamefont {Rice}},\ }\bibfield  {title} {\bibinfo {title} {Monotone smoothing with application to dose-response curves and the assessment of synergism},\ }\href@noop {} {\bibfield  {journal} {\bibinfo  {journal} {Biometrics}\ }\textbf {\bibinfo {volume} {46}},\ \bibinfo {pages} {1071} (\bibinfo {year} {1990})}\BibitemShut {NoStop}%
\bibitem [{\citenamefont {Maatouk}(2022)}]{maatouk2023finite}%
  \BibitemOpen
  \bibfield  {author} {\bibinfo {author} {\bibfnamefont {H.}~\bibnamefont {Maatouk}},\ }\bibfield  {title} {\bibinfo {title} {Finite-dimensional approximation of {G}aussian processes with linear inequality constraints and noisy observations},\ }\href@noop {} {\bibfield  {journal} {\bibinfo  {journal} {Commun. Stat. - Theory Methods}\ }\textbf {\bibinfo {volume} {52}},\ \bibinfo {pages} {8018} (\bibinfo {year} {2022})}\BibitemShut {NoStop}%
\bibitem [{\citenamefont {Riihim{\"a}ki}\ and\ \citenamefont {Vehtari}(2010)}]{riihimaki2010gaussian}%
  \BibitemOpen
  \bibfield  {author} {\bibinfo {author} {\bibfnamefont {J.}~\bibnamefont {Riihim{\"a}ki}}\ and\ \bibinfo {author} {\bibfnamefont {A.}~\bibnamefont {Vehtari}},\ }\bibfield  {title} {\bibinfo {title} {Gaussian processes with monotonicity information},\ }in\ \href@noop {} {\emph {\bibinfo {booktitle} {Proceedings of the thirteenth international conference on artificial intelligence and statistics}}},\ Vol.~\bibinfo {volume} {9}\ (\bibinfo  {publisher} {Proceedings of Machine Learning Research},\ \bibinfo {year} {2010})\ pp.\ \bibinfo {pages} {645--652}\BibitemShut {NoStop}%
\bibitem [{\citenamefont {Maatouk}\ and\ \citenamefont {Bay}(2017)}]{maatouk2017gaussian}%
  \BibitemOpen
  \bibfield  {author} {\bibinfo {author} {\bibfnamefont {H.}~\bibnamefont {Maatouk}}\ and\ \bibinfo {author} {\bibfnamefont {X.}~\bibnamefont {Bay}},\ }\bibfield  {title} {\bibinfo {title} {Gaussian process emulators for computer experiments with inequality constraints},\ }\href@noop {} {\bibfield  {journal} {\bibinfo  {journal} {Math. Geosci.}\ }\textbf {\bibinfo {volume} {49}},\ \bibinfo {pages} {557} (\bibinfo {year} {2017})}\BibitemShut {NoStop}%
\bibitem [{\citenamefont {Jensen}\ \emph {et~al.}(2013)\citenamefont {Jensen}, \citenamefont {Nielsen},\ and\ \citenamefont {Larsen}}]{jensen2013bounded}%
  \BibitemOpen
  \bibfield  {author} {\bibinfo {author} {\bibfnamefont {B.~S.}\ \bibnamefont {Jensen}}, \bibinfo {author} {\bibfnamefont {J.~B.}\ \bibnamefont {Nielsen}},\ and\ \bibinfo {author} {\bibfnamefont {J.}~\bibnamefont {Larsen}},\ }\bibfield  {title} {\bibinfo {title} {Bounded {G}aussian process regression},\ }in\ \href@noop {} {\emph {\bibinfo {booktitle} {2013 IEEE international workshop on machine learning for signal processing (MLSP)}}}\ (\bibinfo {year} {2013})\ pp.\ \bibinfo {pages} {1--6}\BibitemShut {NoStop}%
\bibitem [{\citenamefont {Hertzberg}\ and\ \citenamefont {Mumtaz}(2018)}]{hertzberg2018chemical}%
  \BibitemOpen
  \bibfield  {author} {\bibinfo {author} {\bibfnamefont {R.}~\bibnamefont {Hertzberg}}\ and\ \bibinfo {author} {\bibfnamefont {M.}~\bibnamefont {Mumtaz}},\ }\href@noop {} {\bibinfo {title} {Chemical mixtures and combined chemical and nonchemical stressors}} (\bibinfo {year} {2018})\BibitemShut {NoStop}%
\bibitem [{\citenamefont {Da~Veiga}\ and\ \citenamefont {Marrel}(2012)}]{da2012gaussian}%
  \BibitemOpen
  \bibfield  {author} {\bibinfo {author} {\bibfnamefont {S.}~\bibnamefont {Da~Veiga}}\ and\ \bibinfo {author} {\bibfnamefont {A.}~\bibnamefont {Marrel}},\ }\bibfield  {title} {\bibinfo {title} {Gaussian process modeling with inequality constraints},\ }in\ \href@noop {} {\emph {\bibinfo {booktitle} {Annales de la Facult{\'e} des sciences de Toulouse: Math{\'e}matiques}}},\ Vol.~\bibinfo {volume} {21}\ (\bibinfo {year} {2012})\ pp.\ \bibinfo {pages} {529--555}\BibitemShut {NoStop}%
\bibitem [{\citenamefont {Raissi}\ \emph {et~al.}(2017)\citenamefont {Raissi}, \citenamefont {Perdikaris},\ and\ \citenamefont {Karniadakis}}]{raissi2017machine}%
  \BibitemOpen
  \bibfield  {author} {\bibinfo {author} {\bibfnamefont {M.}~\bibnamefont {Raissi}}, \bibinfo {author} {\bibfnamefont {P.}~\bibnamefont {Perdikaris}},\ and\ \bibinfo {author} {\bibfnamefont {G.~E.}\ \bibnamefont {Karniadakis}},\ }\bibfield  {title} {\bibinfo {title} {Machine learning of linear differential equations using {G}aussian processes},\ }\href@noop {} {\bibfield  {journal} {\bibinfo  {journal} {J. Comput. Phys.}\ }\textbf {\bibinfo {volume} {348}},\ \bibinfo {pages} {683} (\bibinfo {year} {2017})}\BibitemShut {NoStop}%
\bibitem [{\citenamefont {S{\"a}rkk{\"a}}(2011)}]{sarkka2011linear}%
  \BibitemOpen
  \bibfield  {author} {\bibinfo {author} {\bibfnamefont {S.}~\bibnamefont {S{\"a}rkk{\"a}}},\ }\bibfield  {title} {\bibinfo {title} {Linear operators and stochastic partial differential equations in {G}aussian process regression},\ }in\ \href@noop {} {\emph {\bibinfo {booktitle} {Artificial Neural Networks and Machine Learning--ICANN 2011}}},\ Vol.\ \bibinfo {volume} {6792},\ \bibinfo {editor} {edited by\ \bibinfo {editor} {\bibfnamefont {T.}~\bibnamefont {Honkela}}, \bibinfo {editor} {\bibfnamefont {W.}~\bibnamefont {Duch}}, \bibinfo {editor} {\bibfnamefont {M.}~\bibnamefont {Girolami}},\ and\ \bibinfo {editor} {\bibfnamefont {S.}~\bibnamefont {Kaski}}}\ (\bibinfo  {publisher} {Springer},\ \bibinfo {year} {2011})\ pp.\ \bibinfo {pages} {151--158}\BibitemShut {NoStop}%
\bibitem [{\citenamefont {Jidling}\ \emph {et~al.}(2017)\citenamefont {Jidling}, \citenamefont {Wahlstr{\"o}m}, \citenamefont {Wills},\ and\ \citenamefont {Sch{\"o}n}}]{jidling2017linearly}%
  \BibitemOpen
  \bibfield  {author} {\bibinfo {author} {\bibfnamefont {C.}~\bibnamefont {Jidling}}, \bibinfo {author} {\bibfnamefont {N.}~\bibnamefont {Wahlstr{\"o}m}}, \bibinfo {author} {\bibfnamefont {A.}~\bibnamefont {Wills}},\ and\ \bibinfo {author} {\bibfnamefont {T.~B.}\ \bibnamefont {Sch{\"o}n}},\ }\bibfield  {title} {\bibinfo {title} {Linearly constrained {G}aussian processes},\ }in\ \href@noop {} {\emph {\bibinfo {booktitle} {Advances in neural information processing systems}}},\ Vol.~\bibinfo {volume} {30},\ \bibinfo {editor} {edited by\ \bibinfo {editor} {\bibfnamefont {I.}~\bibnamefont {Guyon}}, \bibinfo {editor} {\bibfnamefont {U.~V.}\ \bibnamefont {Luxburg}}, \bibinfo {editor} {\bibfnamefont {S.}~\bibnamefont {Bengio}}, \bibinfo {editor} {\bibfnamefont {H.}~\bibnamefont {Wallach}}, \bibinfo {editor} {\bibfnamefont {R.}~\bibnamefont {Fergus}}, \bibinfo {editor} {\bibfnamefont {S.}~\bibnamefont {Vishwanathan}},\ and\ \bibinfo {editor} {\bibfnamefont {R.}~\bibnamefont {Garnett}}}\ (\bibinfo  {publisher} {Curran
  Associates, Inc.},\ \bibinfo {year} {2017})\BibitemShut {NoStop}%
\bibitem [{\citenamefont {Salzmann}\ and\ \citenamefont {Urtasun}(2010)}]{salzmann2010implicitly}%
  \BibitemOpen
  \bibfield  {author} {\bibinfo {author} {\bibfnamefont {M.}~\bibnamefont {Salzmann}}\ and\ \bibinfo {author} {\bibfnamefont {R.}~\bibnamefont {Urtasun}},\ }\bibfield  {title} {\bibinfo {title} {Implicitly constrained {G}aussian process regression for monocular non-rigid pose estimation},\ }in\ \href@noop {} {\emph {\bibinfo {booktitle} {Advances in neural information processing systems}}},\ Vol.~\bibinfo {volume} {23},\ \bibinfo {editor} {edited by\ \bibinfo {editor} {\bibfnamefont {J.}~\bibnamefont {Lafferty}}, \bibinfo {editor} {\bibfnamefont {C.}~\bibnamefont {Williams}}, \bibinfo {editor} {\bibfnamefont {J.}~\bibnamefont {Shawe-Taylor}}, \bibinfo {editor} {\bibfnamefont {R.}~\bibnamefont {Zemel}},\ and\ \bibinfo {editor} {\bibfnamefont {A.}~\bibnamefont {Culotta}}}\ (\bibinfo  {publisher} {Curran Associates, Inc.},\ \bibinfo {year} {2010})\BibitemShut {NoStop}%
\bibitem [{\citenamefont {Purisha}\ \emph {et~al.}(2019)\citenamefont {Purisha}, \citenamefont {Jidling}, \citenamefont {Wahlstr{\"o}m}, \citenamefont {Sch{\"o}n},\ and\ \citenamefont {S{\"a}rkk{\"a}}}]{purisha2019probabilistic}%
  \BibitemOpen
  \bibfield  {author} {\bibinfo {author} {\bibfnamefont {Z.}~\bibnamefont {Purisha}}, \bibinfo {author} {\bibfnamefont {C.}~\bibnamefont {Jidling}}, \bibinfo {author} {\bibfnamefont {N.}~\bibnamefont {Wahlstr{\"o}m}}, \bibinfo {author} {\bibfnamefont {T.~B.}\ \bibnamefont {Sch{\"o}n}},\ and\ \bibinfo {author} {\bibfnamefont {S.}~\bibnamefont {S{\"a}rkk{\"a}}},\ }\bibfield  {title} {\bibinfo {title} {Probabilistic approach to limited-data computed tomography reconstruction},\ }\href@noop {} {\bibfield  {journal} {\bibinfo  {journal} {Inverse Probl.}\ }\textbf {\bibinfo {volume} {35}},\ \bibinfo {pages} {105004} (\bibinfo {year} {2019})}\BibitemShut {NoStop}%
\bibitem [{\citenamefont {Mie}(1976)}]{mie1976contributions}%
  \BibitemOpen
  \bibfield  {author} {\bibinfo {author} {\bibfnamefont {G.}~\bibnamefont {Mie}},\ }\bibfield  {title} {\bibinfo {title} {Contributions to the optics of turbid media, particularly of colloidal metal solutions},\ }\href@noop {} {\bibfield  {journal} {\bibinfo  {journal} {Ann. Phys.}\ }\textbf {\bibinfo {volume} {25}},\ \bibinfo {pages} {377} (\bibinfo {year} {1976})}\BibitemShut {NoStop}%
\bibitem [{\citenamefont {Seyedheydari}\ \emph {et~al.}(2022)\citenamefont {Seyedheydari}, \citenamefont {Conley}, \citenamefont {Yl{\"a}-Oijala}, \citenamefont {Sihvola},\ and\ \citenamefont {Ala-Nissila}}]{seyedheydari2022electromagnetic}%
  \BibitemOpen
  \bibfield  {author} {\bibinfo {author} {\bibfnamefont {F.}~\bibnamefont {Seyedheydari}}, \bibinfo {author} {\bibfnamefont {K.}~\bibnamefont {Conley}}, \bibinfo {author} {\bibfnamefont {P.}~\bibnamefont {Yl{\"a}-Oijala}}, \bibinfo {author} {\bibfnamefont {A.}~\bibnamefont {Sihvola}},\ and\ \bibinfo {author} {\bibfnamefont {T.}~\bibnamefont {Ala-Nissila}},\ }\bibfield  {title} {\bibinfo {title} {Electromagnetic response and optical properties of anisotropic cusbs2 nanoparticles},\ }\href@noop {} {\bibfield  {journal} {\bibinfo  {journal} {J. Opt. Soc. Am. B}\ }\textbf {\bibinfo {volume} {39}},\ \bibinfo {pages} {1743} (\bibinfo {year} {2022})}\BibitemShut {NoStop}%
\bibitem [{\citenamefont {Vogel}(2002)}]{vogel2002computational}%
  \BibitemOpen
  \bibfield  {author} {\bibinfo {author} {\bibfnamefont {C.~R.}\ \bibnamefont {Vogel}},\ }\href@noop {} {\emph {\bibinfo {title} {Computational methods for inverse problems}}}\ (\bibinfo  {publisher} {SIAM},\ \bibinfo {year} {2002})\BibitemShut {NoStop}%
\bibitem [{\citenamefont {S{\"a}rkk{\"a}}\ and\ \citenamefont {Svensson}(2023)}]{sarkka2023bayesian}%
  \BibitemOpen
  \bibfield  {author} {\bibinfo {author} {\bibfnamefont {S.}~\bibnamefont {S{\"a}rkk{\"a}}}\ and\ \bibinfo {author} {\bibfnamefont {L.}~\bibnamefont {Svensson}},\ }\href@noop {} {\emph {\bibinfo {title} {Bayesian filtering and smoothing}}},\ \bibinfo {series} {The Art of Computer Programming}, Vol.~\bibinfo {volume} {17}\ (\bibinfo  {publisher} {Cambridge university press},\ \bibinfo {year} {2023})\BibitemShut {NoStop}%
\bibitem [{\citenamefont {Conley}\ \emph {et~al.}(2021)\citenamefont {Conley}, \citenamefont {Moosakhani}, \citenamefont {Thakore}, \citenamefont {Ge}, \citenamefont {Lehtonen}, \citenamefont {Karttunen}, \citenamefont {Hannula},\ and\ \citenamefont {Ala-Nissila}}]{conley2021silica}%
  \BibitemOpen
  \bibfield  {author} {\bibinfo {author} {\bibfnamefont {K.}~\bibnamefont {Conley}}, \bibinfo {author} {\bibfnamefont {S.}~\bibnamefont {Moosakhani}}, \bibinfo {author} {\bibfnamefont {V.}~\bibnamefont {Thakore}}, \bibinfo {author} {\bibfnamefont {Y.}~\bibnamefont {Ge}}, \bibinfo {author} {\bibfnamefont {J.}~\bibnamefont {Lehtonen}}, \bibinfo {author} {\bibfnamefont {M.}~\bibnamefont {Karttunen}}, \bibinfo {author} {\bibfnamefont {S.~P.}\ \bibnamefont {Hannula}},\ and\ \bibinfo {author} {\bibfnamefont {T.}~\bibnamefont {Ala-Nissila}},\ }\bibfield  {title} {\bibinfo {title} {Silica-silicon composites for near-infrared reflection: A comprehensive computational and experimental study},\ }\href@noop {} {\bibfield  {journal} {\bibinfo  {journal} {Ceram. Int.}\ }\textbf {\bibinfo {volume} {47}},\ \bibinfo {pages} {16833} (\bibinfo {year} {2021})}\BibitemShut {NoStop}%
\end{thebibliography}%

\end{document}